%% file: main.tex
\documentclass[10pt,twocolumn,letterpaper]{article}

\usepackage{cvpr}              
\usepackage{amsmath}
\usepackage{graphicx}
\graphicspath{ {./Pics/} }
\usepackage{enumerate}
\usepackage{times}
\usepackage{epsfig}
\usepackage{amssymb}
\usepackage{booktabs}
\usepackage{pifont}
\usepackage{bm}
\usepackage{comment}
\usepackage[dvipsnames]{xcolor}
\usepackage{tabularx}
\usepackage[accsupp]{axessibility}
\usepackage{stfloats}
\usepackage{enumitem}
\usepackage{booktabs}
\usepackage{array,multirow}
\usepackage{xspace}
\usepackage{booktabs} 
\usepackage{graphicx} 
\usepackage{soul}
\usepackage{xcolor}
\usepackage[theorems,skins,breakable]{tcolorbox}
\usepackage[normalem]{ulem}
\usepackage{float} 
\definecolor{mypurple}{RGB}{128,0,128} 
\definecolor{mygreen}{RGB}{0,128,0} 
\definecolor{mydarkblue}{RGB}{6,57,112} 
\definecolor{mylightblue}{RGB}{3, 165, 252}
\definecolor{hlightyellow}{RGB}{255,255,204}
\definecolor{lightyellow}{RGB}{255,255,204}

\newtcolorbox[list inside=template,auto counter,number within=section]{template}[1][]{
    colback=hlightyellow,
    #1, 
}

\newtcolorbox[list inside=prompt,auto counter,number within=section]{prompt}[1][]{
    enhanced,breakable,
    colbacktitle=black!60,
    coltitle=white,
    fontupper=\small,
    boxsep=5pt,
    left=0pt,
    right=0pt,
    top=0pt,
    bottom=0pt,
    boxrule=1pt,
    title={#1},
    #1, 
}

\newcommand{\hrt}[1][red]{\textcolor{#1}}

\newcommand{\Simin}[1][black]{\textcolor{#1}}

\definecolor{cvprblue}{rgb}{0.21,0.49,0.74}
\usepackage[pagebackref,breaklinks,colorlinks,citecolor=cvprblue]{hyperref}
\def\paperID{4800} 
\def\confName{CVPR}
\def\confYear{2024}

\title{JRDB-Social: A Multifaceted Robotic Dataset for Understanding of Context and Dynamics of Human Interactions Within Social Groups}

\author{Simindokht Jahangard, Zhixi Cai, Shiki Wen, Hamid Rezatofighi\\
Monash University\\
\small \texttt{\{simindokht.jahangard,zhixi.cai,hamid.rezatofighi\}@monash.edu,swen0021@student.monash.edu}
\\
\url{https://jrdb.erc.monash.edu/dataset/social}
}
\begin{document}
\twocolumn[{
\maketitle
\begin{center}
\vspace{-2em}
    \captionsetup{type=figure}
\includegraphics[width=1\textwidth]{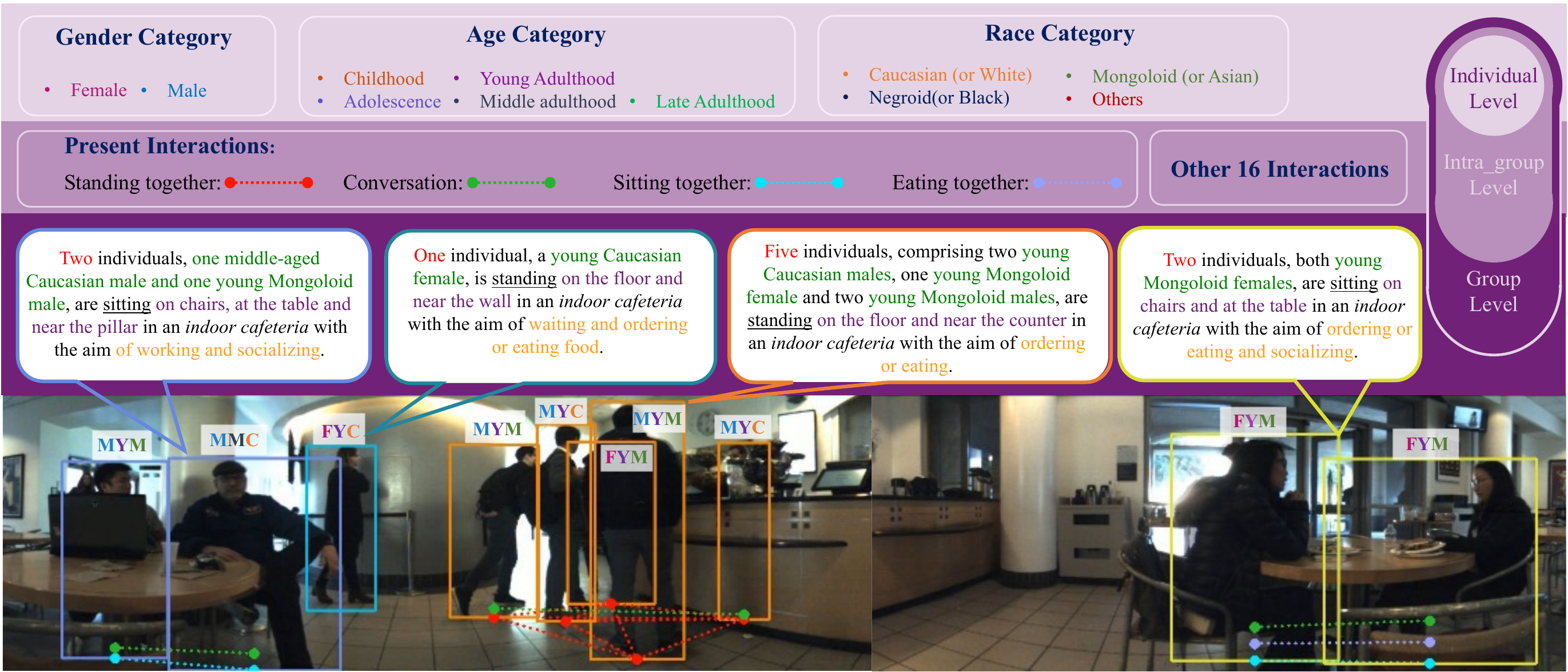}\vspace{-.5em}
    \captionof{figure}{Some highlighted instances from the JRDB-Social dataset featuring detailed annotations across three levels: {\bf Individual Level)} Representing specific attributes like age, gender, and race are shown through color-coded abbreviations. For example, `\textcolor{mylightblue}{M}\textcolor{mydarkblue}{M}\textcolor{orange}{C}' represents \textcolor{mylightblue}{Male}, \textcolor{mydarkblue}{Middle Adulthood}, \textcolor{orange}{Caucasian}. {\bf Intra-group Level)} This level focuses on group dynamics and interactions between each pair at the frame level, represented by dashed lines. {\bf Group Level)} Each social group~\cite{ehsanpour2020joint} is represented by the same colour and accompanied by textual descriptions that detail the \hrt{ number of members}, \textcolor{mygreen}{their specific attributes}, their \textcolor{mypurple}{body position’s connection with the content}, the presence of \textcolor{mypurple}{salient scene content} near the group, the \textit{venue}, and the \textcolor{orange}{group's aim or purpose}.} 
\label{fig:tisser}
\end{center}
}]

\input{sec/0_abstract}    
\input{sec/1_intro}
\input{sec/2_related-works}
\input{sec/3_Dataset}

\input{sec/3_Evaluation}

\input{sec/4_Experiment}
\input{sec/5_Conclusion}

{
    \clearpage
    \small
    \bibliographystyle{unsrtnat}
    \bibliography{main}
}

\maketitlesupplementary


\input{sec_sup/1_dataset}
\input{sec_sup/2_experiments}


\end{document}

%% file: sec/0_abstract.tex
\begin{abstract}\vspace{-2em}
\noindent Understanding human social behaviour is crucial in computer vision and robotics. Micro-level observations like individual actions fall short, necessitating a comprehensive approach that considers individual behaviour, intra-group dynamics, and social group levels for a thorough understanding.
To address dataset limitations, this paper introduces JRDB-Social, an extension of JRDB~\cite{martin2021jrdb}. Designed to fill gaps in human understanding across diverse indoor and outdoor social contexts, JRDB-Social provides annotations at three levels: individual attributes, intra-group interactions, and social group context. This dataset aims to enhance our grasp of human social dynamics for robotic applications.
Utilizing the recent cutting-edge multi-modal large language models, we evaluated our benchmark to explore their capacity to decipher social human behaviour. 
\end{abstract}
\vspace{-2em}

%% file: sec/1_intro.tex
\section{Introduction}
Human social behaviour understanding finds numerous applications in computer vision and robotics. Simply observing the micro-level information like the actions of an individual is inadequate for a comprehensive understanding of human behaviour because humans are inherently social beings and require analysis within a broader social context. Therefore, a comprehensive and multi-layered approach is required to perceive human social behaviour thoroughly. 
For example, in security and surveillance systems, integrating individual-level data, identifying social groups, and taking context into account significantly enhance the overall capacity to better understand crowd behaviors~\cite{gong2011security}. Additionally, this integration fosters more natural and intuitive experiences in human-robot interaction like telerobots~\cite{barua2020can}, coworker robots~\cite{hanheide2017and} and social robots~\cite{logan2019social}.\\
In recent years, significant progress has been made in vision-based understanding of human behaviour and activity, furnishing datasets at different levels. Some datasets focused on individual-level information such as human attributes and atomic actions~\cite{yatskar2016situation,tanisik2016facial,tanisik2021multi,ronchi2015describing,ryoo2010ut,lee2022human,patron2010high,marszalek2009actions}. 
Conversely, other datasets primarily concentrate on human-human interactions~\cite{tanisik2021multi,ronchi2015describing,ryoo2010ut,lee2022human,patron2010high,marszalek2009actions}. On a higher level, certain datasets provide information regarding human groups and video captioning, describing various events occurring in videos~\cite{alameda2015salsa,wang2020panda,ehsanpour2022jrdb,krishna2017dense,huang2020multimodal,zhou2018towards,tan2020learning,wang2019vatex}.
While serving as valuable resources for the research community, these datasets mainly consider one aspect of this multi-level hierarchy in understanding human behaviour and activity and fall short in adequately capturing and reflecting the complexity of dynamics and context inherent in human social behaviours within crowded scenes.
To bridge this gap, we introduce JRDB-Social an extension of the JRDB dataset~\cite{martin2021jrdb}. JRDB features a social manipulator robot with stereo RGB 360° cameras, dual LiDAR sensors for 3D point clouds, audio, GPS, and over 1.8 million annotations in the form of 2D bounding boxes and 3D oriented cuboids.
The JRDB dataset has already contained very useful annotations such as human atomic actions and social grouping~\cite{ehsanpour2022jrdb} and human body pose annotations~\cite{vendrow2023jrdb}. Our proposed annotations serve as a perfect complement to enrich this popular dataset. JRDB-Social is structured at three distinct levels including: {\bf individual level}, {\bf intra-group level} and the {\bf social group level}. Firstly, at the individual level, we provide annotations for gender, age, and race. Secondly, at the intra-group level, we capture fine-grained, dynamic multi-interactions between (20 categories) each pair within a sub-group at the frame level.
Lastly, at the social group level, we incorporate text captions that describe information about the context including the connection between the group's body position and the content, the presence of salient scene content situated in close proximity to the group, the specific location or venue, and the group's aim and purpose,
thus offering a holistic contextual overview.
\Simin{This benchmark facilitates exploration into how demographic factors influence social behaviour, allowing for examination of differences in interactions based on gender or race. Venue annotations provide contextual information for interactions, recognizing that behaviours and social dynamics in settings like cafeterias may differ from those in formal environments such as classrooms. Understanding the purpose of a group can illuminate the motivation behind the interaction, whether the group gathers for leisure, work, or education. Ultimately, this benchmark seeks to narrow gaps in comprehending human behavior within social settings, furnishing valuable insights to enrich our understanding of social dynamics.}
\\
\Simin{With the surge in popularity and significant advancements in large language models (LLMs) and vision-language models (VLMs)~\cite{zhang2023video, wu2023next, maaz2023video, li2023videochat}, which claim proficiency in visual understanding and reasoning, we explore and assess their capabilities using our dataset. We applied these models to our dataset to evaluate their effectiveness in perceiving and reasoning about human social behavior in crowded environments. Our evaluation focuses on examining and discussing the strengths and limitations of current methodologies in understanding human social and contextual interaction dynamics.} \\
In sum, the key contributions of this work are as follows:
\begin{itemize}
\item Providing JRDB-Social benchmark on dynamic human-human interactions at the frame level, revealing multi-label annotations between each pair within a group.
\item We offer individual attribute annotation and descriptions of social groups. These descriptions elaborate on the relationship between the group's body position and the content, the presence of salient scene context near the group, the venue location, and the group's aim or purpose.
\item We assess the performance of the most recent vision-language models within the framework of JRDB-Social, performing a comprehensive examination to identify the advantages and shortcomings of current approaches.
\end{itemize}

%% file: sec/2_related-works.tex
\section{Related works}
\subsection{Datasets}
 In the following section, we provide the commonly used public datasets across three distinct levels \ie individual, intra-group, and social group level.\\
\textbf{Individual Level.}
Analysing individual-level human behaviour, which encompasses factors like age, gender, and race, alongside detailed atomic action data, is paramount across diverse domains. 
The MovieGraph dataset~\cite{vicol2018moviegraphs} specializes in delineating inferred properties of human-centric situations through intricate, graph-based annotations of social scenarios depicted in movie clips.
Also, recently, autonomous vehicle datasets like~\cite{rasouli2017ICCVW,rasouli2019pie} have been released featuring individual-level annotations comprehending the behaviour of various age groups and genders in traffic scenarios.
Conversely, certain datasets, such as~\cite{gu2018ava,yun2012two,shahroudy2016ntu,mlbcaptions2018}, focus on atomic actions by offering comprehensive data that specifically highlights individual actions within their content.
Shifting focus, other datasets delve into emotions~\cite{luo2020arbee}, providing additional layers of information to understand human behaviour by considering variables such as age, gender, and ethnicity. 
While valuable, existing datasets lack a perspective from the robot within a social environment and they are not from human crowded environments. JRDB-Social addresses this gap by providing demographic information in real-world data from the robot's perspective.\\
\textbf{Intra-Group Level Interactions.} Some image-based datasets focus intensely on specific interactions such as~\cite{yatskar2016situation,tanisik2016facial,tanisik2021multi,ronchi2015describing,cui2021s,orcesi2021detecting}. Also, some video-based including~\cite{ryoo2010ut,patron2010high,marszalek2009actions,alameda2015salsa,lee2022human,yun2012two} offer a diverse range of interaction scenarios, contributing to understanding of human interaction dynamics in various contexts.
 The drawback of these datasets lies in their limited number of label categories or the treatment of interaction labels as a subset. Moreover, they often involve interactions between only two or very few individuals, lacking representation of crowd dynamics. JRDB-Social offers frame-level multi-label annotation of human interactions within social groups in crowded scenes.\\
\textbf{Social Group Level.}
A more comprehensive understanding of human behaviour emerges when contextual information is available. In this context, certain datasets provide higher-level information such as~\cite{alameda2015salsa,wang2020panda,ehsanpour2022jrdb} furnish valuable insights into social group dynamics.
On the other hand, some datasets such as~\cite{krishna2017dense,huang2020multimodal,zhou2018towards,tan2020learning,wang2019vatex} that primarily focus on video captioning, offer sets of descriptions for multiple events occurring in videos and aim to temporally localize them.
However, they often overlook crucial, detailed information—especially pertaining to how individuals interact with each other and their surroundings. JRDB-Social offers comprehensive information by providing group-level details such as the group's body position related to the content, the proximity of salient scene content within the group, the group's objective, and key information about the main environment. This approach enhances human understanding by presenting a more holistic view of the scenario.
\subsection{Vision-based Large Language Models} 
In recent years, Large Language Models (LLMs)~\cite{chowdhery2022palm,bai2022constitutional,ouyang2022training,radford2019language} have made significant strides in achieving multi-modal capabilities. Notable models include Video-LLaMA~\cite{zhang2023video}, which enhances LLMs for detailed video comprehension, and NExT-GPT~\cite{wu2023next}, a holistic multi-modal model navigating text, images, videos, and audio seamlessly. Other models like VideoChat~\cite{li2023videochat}, Visual ChatGPT~\cite{wu2023visual}, VALLEY~\cite{luo2023valley}, Otter~\cite{li2023otter}, ViperGPT~\cite{suris2023vipergpt}, and MiniGPT-4~\cite{zhu2023minigpt} contribute to advancements in video understanding, visual processing, and instruction tuning for improved contextual learning. Additionally, efforts such as InstructBLIP~\cite{instructblip}, M$^{3}$IT~\cite{li2023m}, and VisionLLM~\cite{wang2023visionllm} focus on instruction tuning, multilingual datasets, and vision-centric tasks, collectively propelling AI systems towards greater versatility in language understanding and nuanced video comprehension.\\
While these models excel in understanding and reasoning over videos, their capacity to comprehend human social behaviour and conduct contextual activity analysis remains unexplored. This paper aims to assess their performance on the JRDB-Social dataset.

%% file: sec/3_Dataset.tex
\section{The JRDB-Social Dataset}
We developed JRDB-Social to complement the current annotation of JRDB dataset~\cite{martin2021jrdb} by providing new annotations to better comprehend human activity in a social context. JRDB dataset contains 64 minutes of sensory data, comprising 54 sequences reflecting diverse indoor and outdoor locations within a university campus environment. The JRDB dataset has been captured by a social manipulator robot featuring stereo RGB 360° cameras, dual LiDAR sensors for 3D point clouds, audio, GPS, and boasts over 1.8 million annotations in the form of 2D bounding boxes and 3D oriented cuboids. The JRDB dataset already contains valuable annotations, such as human atomic actions, social grouping~\cite{ehsanpour2022jrdb}, and human body pose annotations~\cite{vendrow2023jrdb} and JRDB-Social serves as a complementary extension to this dataset, providing a multifaceted perspective at three levels: the individual, intra-group, and the social group level. \\
\textbf{Annotation Process.}
\Simin{For annotating JRDB-Social, at each level, we designed a toolbox, featuring unique IDs corresponding to existing 2D and 3D bounding box annotations. 
We adhered to a quality assessment protocol aligned with established benchmarks known for high-quality annotated data, such as previous JRDB benchmarks~\cite{martin2021jrdb,ehsanpour2022jrdb,vendrow2023jrdb}. In line with these benchmarks, we implemented a standardised data annotation process to ensure consistency with past JRDB annotations~\cite{martin2021jrdb,ehsanpour2022jrdb}; for instance, our interaction annotations align seamlessly with the actions of each individual involved. Also, our annotators, chosen for their expertise in behaviour analysis, adhere to strict guidelines and protocols for standardized annotation. Encountering challenges such as significant distance from the robot, varying lighting conditions, occlusion, and crowded scenes, each label in our dataset is accompanied by difficulty level—categorized as \textit{Easy} (1), \textit{Medium} (2), or \textit{Hard} (3)—reflecting the annotator's confidence. To ensure fairness and consistency, labels undergo a thorough review by two additional individuals, alongside random quality assessments by multiple assessors.}\\
\textbf{Text Description Structure.}  
We enhance JRDB-Social by including text descriptions for each group to offer contextual understanding. This aligns with the trend of combining natural language understanding with computer vision, benefiting tasks like image captioning. This enhancement also has potential in Human-Robot interaction, helping robots adjust behaviour based on group context, thus improving interactions. We construct our sentences in the colour-coded format, shown in the yellow box below. 
\begin{template}[float, floatplacement=t]
    \textcolor{black}{{\bf Text Description Structure:}\hrt{[number of individuals]}, including attribute of each person involved (e.g., Person 1:\textcolor{mygreen}{[age, gender, race]}, Person 2:\textcolor{mygreen}{[age, gender, race]}, Person 3:\textcolor{mygreen}{[age, gender, race]}, and so on). These individuals engage in activities on \textcolor{mypurple}{[the content relates to group's body position and the presence of a salient scene content nearby]} in \textit{[a specific venue location]} with the purpose of \textcolor{orange}{[group's goal]}}.
\end{template}
\subsection{Individual Level Attributes}
\begin{figure}[b]
\vspace{-1em}
\begin{center}
\scalebox{1}{
 \includegraphics[width=1\linewidth]{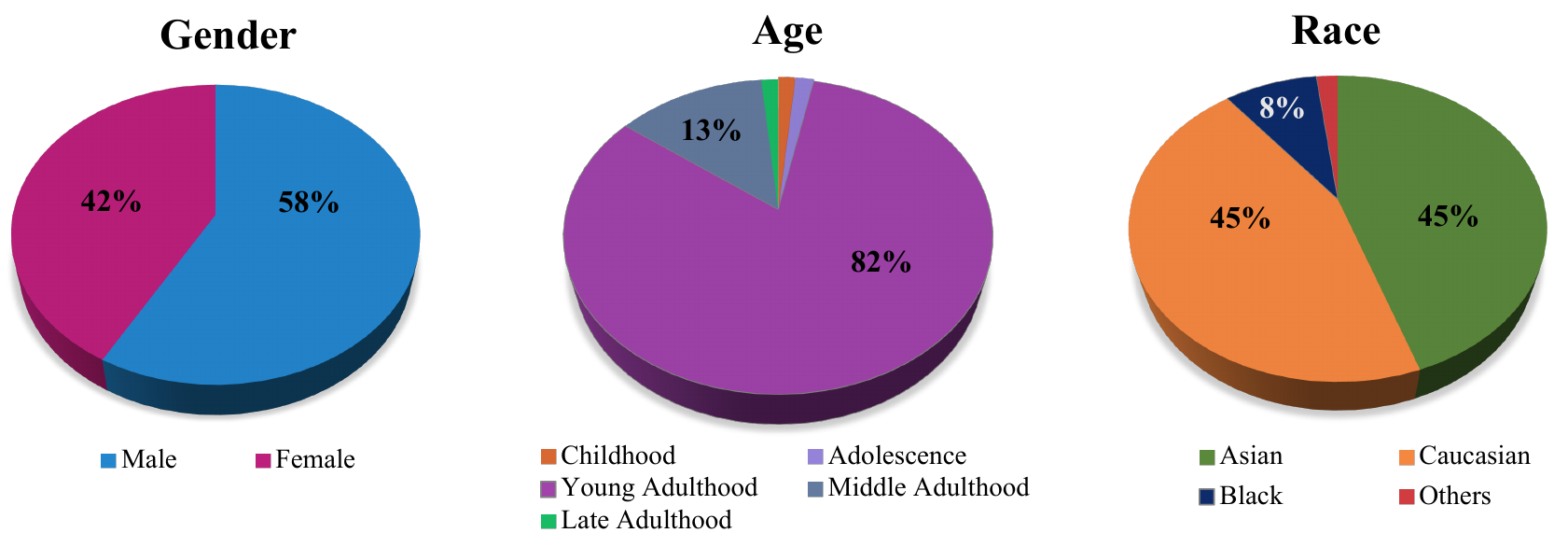}
}
\end{center}
\vspace{-2em}
  \caption{Statistics of individual attributes.}
\label{fig:GAE}
\end{figure}

The JRDB-Social dataset includes individual attributes, as understanding these is crucial for studying diverse social behaviours in groups and deepening insights into human behaviour in social situations especially in social sciences and psychology research. Additionally, in human-robot interactions, awareness of individuals' demographics aids in personalizing the robot's behaviour for more culturally sensitive interactions.
Therefore, in addition to the currently available annotations of human body pose and atomic action in~\cite{vendrow2023jrdb, ehsanpour2022jrdb}, we annotated gender, age, and race in this dataset. Under the gender category, the dataset distinctly classifies individuals into two primary groups: \textit{Male} and \textit{Female}. The age attribute is finely segmented into five distinct groups: \textit{Childhood} (3-12 years), \textit{Adolescence} (13-20 years), \textit{Young Adulthood} (21-40 years), \textit{Middle Adulthood} (41-65 years), and \textit{Late Adulthood} (66 years and above). In terms of racial classification, the dataset adopts Alfred L. Kroeber's classification\footnote{\url{https://en.m.wikipedia.org/wiki/Mongoloid}} which is based on physical characteristics. It includes \textit{Caucasian/White} (light skin, varied eye colours), \textit{Negroid/Black} (dark skin, coiled hair), \textit{Mongoloid/Asian} (almond-shaped eyes, black hair, varied skin tones). Figure~\ref{fig:GAE} illustrates attribute distributions within the JRDB-Social dataset excluding impossible ones. As illustrated, male individuals predominate in the gender category. The video, primarily captured in a university environment, predominantly features individuals in the young adulthood category, reflecting distribution of this category in real-life situations. The racial breakdown shows equal representation from Caucasian and Asian populations, with a smaller proportion representing the Black community.
Figure~\ref{fig:tisser} shows some samples.
\subsection{Intra-Group Level Dynamic Interactions}
\begin{figure}[t]
\begin{center}
\scalebox{0.99}{
 \includegraphics[width=1\linewidth]{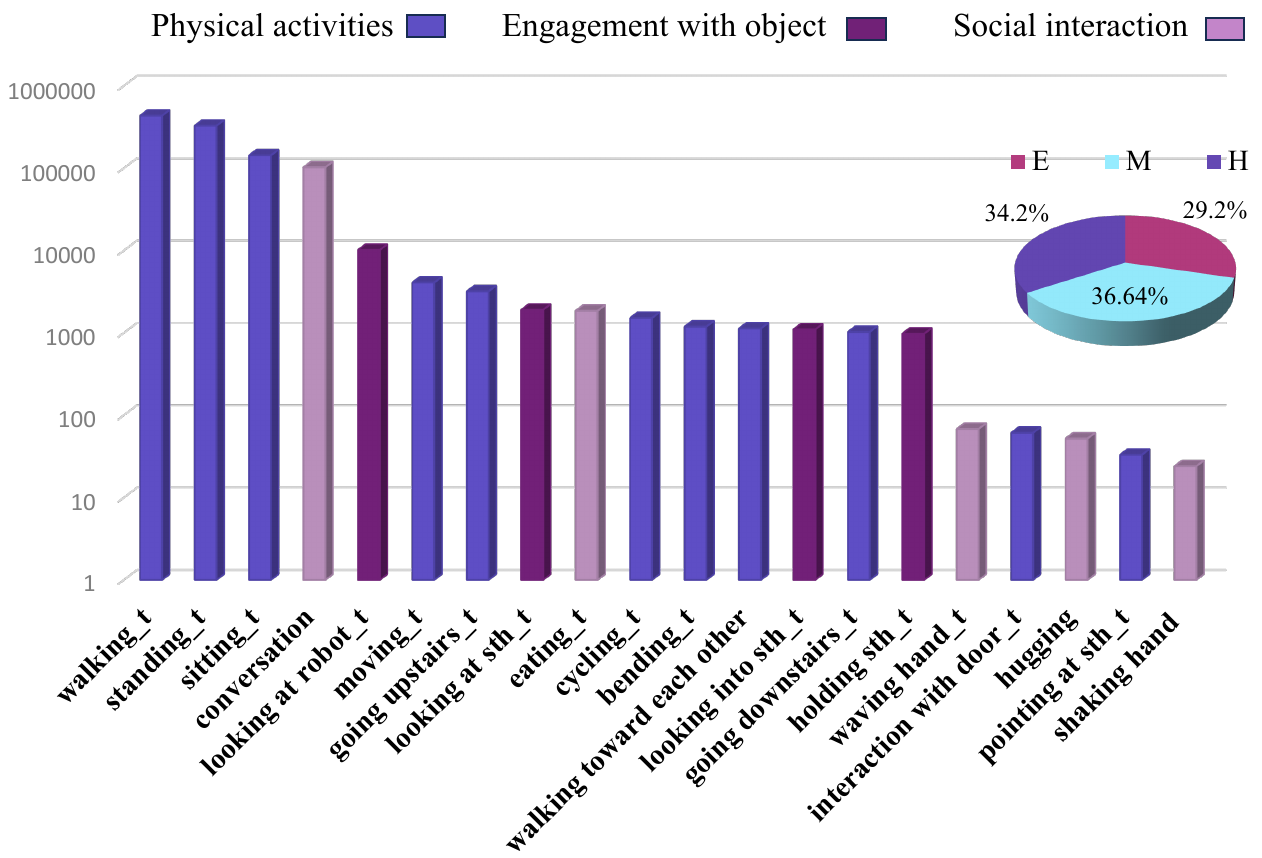}}
\end{center}\vspace{-2em}
  \caption{Sorting interaction classes on a log-scale distribution, displaying descending frame numbers for all data. Difficulty levels indicated as E (Easy), M (Medium), and H (Hard).}
\label{fig:Interaction}
\vspace{-1em}
\end{figure}

The concept of multi-label interaction at the frame level provides a detailed understanding of social dynamics within social groups~\cite{ehsanpour2022jrdb}, offering detailed insights into simultaneous actions and gestures among individuals. These fine-grained annotations are instrumental in training machine learning models for the recognition of diverse social interactions, especially in social navigation scenarios. Additionally, the frame-level annotations facilitate behavioural studies, allowing researchers to examine in-depth the temporal dynamics of interactions and how individuals engage with each other in specific social settings. In JRDB-Social, we provided multi-label fine-grained interaction annotation at the frame level and categorized it into three distinct groups, each encompassing various dimensions of shared experiences. The first category, shown in purple in Figure \ref{fig:Interaction}, focuses on shared physical activities, including behaviours with physical proximity and posture. The second, in dark purple, involves joint engagement with external entities, often centred around interacting with objects together. The third, in light pink, encompasses interpersonal exchanges and gestures as part of social interactions.
The distribution of dynamic interaction classes for both training and test sets is depicted in Figure~\ref{fig:Interaction}. The vertical axis, presented on a log scale, represents the number of frames. Notably, prevalent interactions include walking, standing, sitting together, and engaging in conversations and less frequent activities like pointing at something together and shaking hands well reflect distribution biases in real-world daily scenarios. Additionally, the accompanying pie chart illustrates difficulty levels, with medium difficulty comprising the largest portion at 36.64\%, followed by hard at 34.2\%, and easy at 29.2\%, indicating an even distribution of difficulty. 
During the annotation process, interactions between individuals within each group are meticulously annotated. We identify the individuals participating, document the frame range of the interaction, and to improve accuracy, integrate the individual actions outlined in~\cite{ehsanpour2022jrdb}, aligning them with the corresponding interaction. More details about our protocols are provided in supplementary materials.
\subsection{Social Group Level Context}
\begin{figure}[t]
\begin{center}
\scalebox{0.99}{
 \includegraphics[width=1\linewidth]{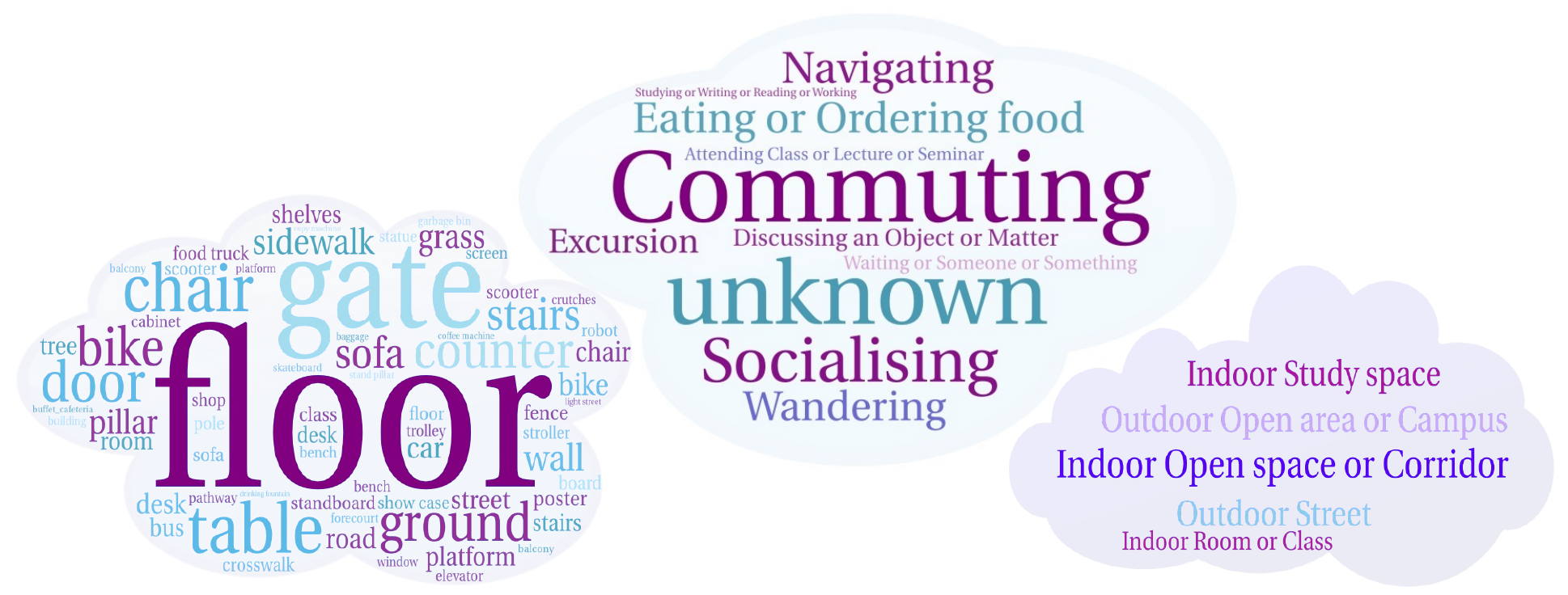}
}
\end{center}\vspace{-1em}
  \caption{Social group level word cloud in the dataset. Left: location of body posture and objects. Top: group aim. Right: venue locations. Larger words indicate higher frequency.
}
\label{fig:intention}
\vspace{-1em}
\end{figure}

These annotations aim to provide a comprehensive understanding of social behaviour at the social group level. By including information beyond individual attributes and interactions, the dataset becomes richer and more reflective of real-world scenarios involving groups of people and context.
It encompasses the details about the group's surrounding environment, their specific venue and aims and purposes.
Figure~\ref{fig:intention} illustrates a word cloud depicting labels for each category. Further statistical details for each category can be found in the supplementary materials.\\
\textbf{Engagement of Body Position with the Content and Salient Scene Content.} 
These annotations contribute to a contextual analysis of the physical engagement of the group and add layers of context to the dataset. This involves the examination of body position related to the content (BPC), considering the majority of group members. Additionally, it offers valuable insights into the presence of salient scene elements (SSC) in their surroundings.
To elaborate, the BPC encompasses specifics regarding how body position is linked to the content. For instance, sitting on a \textit{chair} or standing on the \textit{platform}. On the other hand, SSC provides information about the presence of dominant scene content near the group. This includes observations like standing near a \textit{pillar} or \textit{counter}. As the location of the group may vary, we annotate this information at the frame level.\\
\textbf{The Venue Location.}
This annotation offers information about the locations where the group participates in activities, helping in modelling and predicting the movement patterns of individuals and groups. This is essential for robots to navigate through diverse environments, adapting their behaviour based on the spatial context. These are classified into indoor spaces like \textit{cafeterias, dining halls, or food courts}, \textit{open spaces or corridors}, \textit{rooms or classrooms}, and \textit{study areas}. Furthermore, it includes outdoor categories such as \textit{open areas or campuses} and \textit{streets}.\\
\textbf{Group's Aims and Purposes.} 
These annotations provide information at the social group level about the purpose behind the formation and activities of each group, the dataset becomes a valuable resource for advancing research in social understanding, behavioural analysis, and contextual reasoning.
Our categorization provides information from the act of moving through spaces, utilizing corridors in \textit{navigating}, to routine travel in \textit{commuting}, and aimless strolls in \textit{wandering}, the categories capture various facets of human interaction. \textit{Socializing} emphasizes communal connections, while \textit{studying, writing, reading, and working} highlighting focused intellectual activities. \textit{Discussing an object or a matter} centres on engaging conversations around specific topics, and \textit{attending class, lecture, or seminar} underscores educational gatherings. \textit{Ordering and eating food} portrays communal aspects of meal-related activities, and \textit{excursion} adds a recreational dimension to the group's aim. Moreover, \textit{Waiting for someone or something} demonstrates the anticipation and patience associated with awaiting a person or an event. In essence, this categorization offers nuanced insights into the multifaceted dynamics of collective human behaviour in diverse contexts.

%% file: sec/3_Evaluation.tex
\begin{table*}[t]
   \centering
   \footnotesize
   \begin{tabular}{lccc|c|cccc|c}
    \toprule
    \multirow{2}{*}{{\bf Multi-modal LLM}} & \multicolumn{3}{c}{{\bf Individual Level}} & \multicolumn{1}{c}{{\bf Intra-Group Level}} & \multicolumn{4}{c}{{\bf Social Group Level}} & \textbf{Overall} \\
    \cline{2-10}
    & Gender & Age & Race & Interactions & BPC & SSC & Venue & Purpose & Average \\
    \hline
    Video-LLaMA (LLaMA-2 13B)~\cite{zhang2023video} & \underline{0.7139} & \underline{0.3069} & {\bf0.2837} & \underline{0.3253} & 0.1639 & 0.1252 & {\bf0.2413} & 0.2595 & \textbf{0.3025} \\
    Video-LLaMA (LLaMA-2 7B)~\cite{zhang2023video} & 0.5200 & {\bf0.3196} & 0.2308 & {\bf0.3852} & \underline{0.1642}& 0.1639 & \underline{0.2147} & {\bf0.3003} & \underline{0.2874} \\
    Valley (LLaMA-1 13B)~\cite{luo2023valley} & 0.3991 & 0.2041 & 0.1253 & 0.1674 & 0.0364 & 0.0456 & 0.0904 & 0.2632 & 0.1603 \\
    Valley (LLaMA-2 7B)~\cite{luo2023valley} & 0.4658 & 0.1731 & 0.0905 & 0.2035 & 0.1115 & 0.0559 & 0.0695 & 0.2515 & 0.1400 \\
    OTTER (LLaMA-1 7B)~\cite{li2023otter} & 0.1959 & 0.1131 & 0.0115& 0.2761 & 0.0799 & 0.1242 & 0.0420 &0.0411 & 0.1105 \\
    MiniGPT-4 (LLaMA-2 7B)~\cite{zhu2023minigpt} & {\bf0.7391} & 0.2204 & \underline{0.2604} & 0.2068 & {\bf0.1970} & 0.0978 & 0.2574 & \underline{0.2736} & 0.2816 \\
    InstructBLIP (Vicuna-V1 13B)~\cite{instructblip} & 0.5860 & 0.2482 & 0.1875 & 0.0665& 0.0639 & {\bf0.2354} & 0.1636 & 0.1841 & 0.2169 \\
    InstructBLIP (Vicuna-V1 7B)~\cite{instructblip} & 0.6444 & 0.0697 & 0.2587 & 0.0937 & 0.1337 &0.1174 & 0.2020 & 0.1880 & 0.2135\\
    \hline
  \end{tabular}
\vspace{-1em}
\caption{{\bf Guided Perception} Experiment: Comparing popular multi-modal LLMs across the JRDB-Social in F1 score for all sets. Optimal results in bold, second best underlined. BPC = Engagement of Body Position’s connection with the Content, SSC = Salient Scene Content.}
\label{table:BBf1}
\end{table*}

\begin{table*}[t]
   \centering
   \footnotesize
   \begin{tabular}{lccc|c|cccc|c}
    \toprule
    \multirow{2}{*}{{\bf Multi-modal LLM}} & \multicolumn{3}{c}{{\bf Individual Level}} & \multicolumn{1}{c}{{\bf Intra-Group Level}} & \multicolumn{4}{c}{{\bf Social Group Level}} & \textbf{Overall} \\
    \cline{2-10}
    & Gender & Age & Race & Interactions & BPC & SSC & Venue & Purpose & Average \\
    \hline
    Video-LLaMA (LLaMA-2 13B)~\cite{zhang2023video} & {\bf0.3338} & {\bf0.2543} & {\bf0.3507} & \underline{0.2786}& \underline{0.0795} & 0.0238 & \underline{0.2471} & \underline{0.1792} & \textbf{0.1965} \\
    Video-LLaMA (LLaMA-2 7B)~\cite{zhang2023video} &0.2177 & \underline{0.2256} & \underline{0.2984} & {\bf0.2970} & 0.0637 & 0.0195& {\bf0.2705} & 0.1375 & 0.1645 \\
    Valley (LLaMA-2 7B)~\cite{luo2023valley} & 0.0215 & 0.0579 & 0.0122 & 0.0104 & 0.0025 & 0.0008 & 0.0449 & 0.0211 & 0.0217 \\
    MiniGPT-4 (LLaMA-2 7B)~\cite{zhu2023minigpt} & \underline{0.2344} & 0.2109 & 0.2619 & 0.0994 & {\bf0.0829}& \underline{0.0282} & 0.2432 & {\bf0.1861} & \underline{0.1684} \\
    InstructBLIP (Vicuna-V1 13B)~\cite{instructblip} & 0.0856 & 0.0856 & 0.0346 & 0.0542 & 0.0172 & 0.0267 & 0.1119 & 0.0778 & 0.0643 \\
    InstructBLIP (Vicuna-V1 7B)~\cite{instructblip} & 0.1111 & 0.1478 & 0.0686 & 0.0841 & 0.0314 & {\bf0.0338} & 0.1457 & 0.0821 & 0.0881 \\
    \hline
  \end{tabular}
\vspace{-1em}
\caption{
{\bf Holistic (Counting) Experiment}: Comparing popular multi-modal LLMs across the JRDB-Social in F1 score for all sets. Optimal results in bold, second best underlined. BPC = Engagement of Body Position’s connection with the Content, SSC = Salient Scene Content.}
\label{table:Holif1_C}
\end{table*}

\begin{table*}[t]
   \centering
   \footnotesize
   \begin{tabular}{lccc|c|cccc|c}
    \toprule
    \multirow{2}{*}{{\bf Multi-modal LLM}} & \multicolumn{3}{c}{{\bf Individual Level}} & \multicolumn{1}{c}{{\bf Intra-Group Level}} & \multicolumn{4}{c}{{\bf Social Group Level}} & \textbf{Overall} \\
    \cline{2-10}
    & Gender & Age & Race & Interactions & BPC & SSC & Venue & Purpose & Average \\
    \hline
    Video-LLaMA (LLaMA-2 13B)~\cite{zhang2023video} & {\bf0.9800} & \underline{0.5657} & 0.7458 & 0.2786& 0.2326 & \underline{0.0771} & 0.2814 & \underline{0.4788} & 0.4622 \\
    Video-LLaMA (LLaMA-2 7B)~\cite{zhang2023video} &\underline{0.9800} & 0.5633 & \underline{0.7482} & \underline{0.3213} & 0.2177 & 0.0730& 0.2810 & 0.4663 & 0.4564 \\
    Valley (LLaMA-2 7B)~\cite{luo2023valley} & 0.6958 & 0.4415 & 0.1629 & 0.2602 & 0.2298 & 0.0637 & \underline{0.3350} & 0.4415 & 0.3288\\
    OTTER (LLaMA-1 7B)~\cite{li2023otter} & 0.8194 & 0.5290 & 0.5687 & 0.3053& 0.2796 &0.0913 & {\bf0.4309} & 0.3198 & 0.4542 \\
    MiniGPT-4 (LLaMA-2 7B)~\cite{zhu2023minigpt} & 0.8194 & 0.5290 & 0.5687 & 0.2796 & 0.0913& 0.0282 & {\bf0.4309} & 0.3198 & 0.4180 \\
    InstructBLIP (Vicuna-V1 13B)~\cite{instructblip} & 0.8493 & {\bf0.6223} & {\bf0.7663} & {\bf0.3318} & {\bf0.4045} & {\bf0.1026} & 0.3197 & {\bf0.4797} & \underline{0.4846} \\
    InstructBLIP (Vicuna-V1 7B)~\cite{instructblip} & 0.7621 & 0.5408 & 0.6450 & 0.3081 &\underline{0.2947} & 0.0711 &0.3313 & 0.4326 & \textbf{0.4848} \\
    \hline
  \end{tabular}
\vspace{-1em}
\caption{
{\bf Holistic (Binary) Experiment}: Comparing popular multi-modal LLMs across the JRDB-Social in F1 score for all sets. Optimal results in bold, second best underlined. BPC = Engagement of Body Position’s connection with the Content, SSC = Salient Scene Content.}
\label{table:Holif1_B}
\end{table*}

%% file: sec/4_Experiment.tex
\section{Experiments}
\begin{figure}
\begin{center}
\scalebox{1}{
\includegraphics[width=1\linewidth]{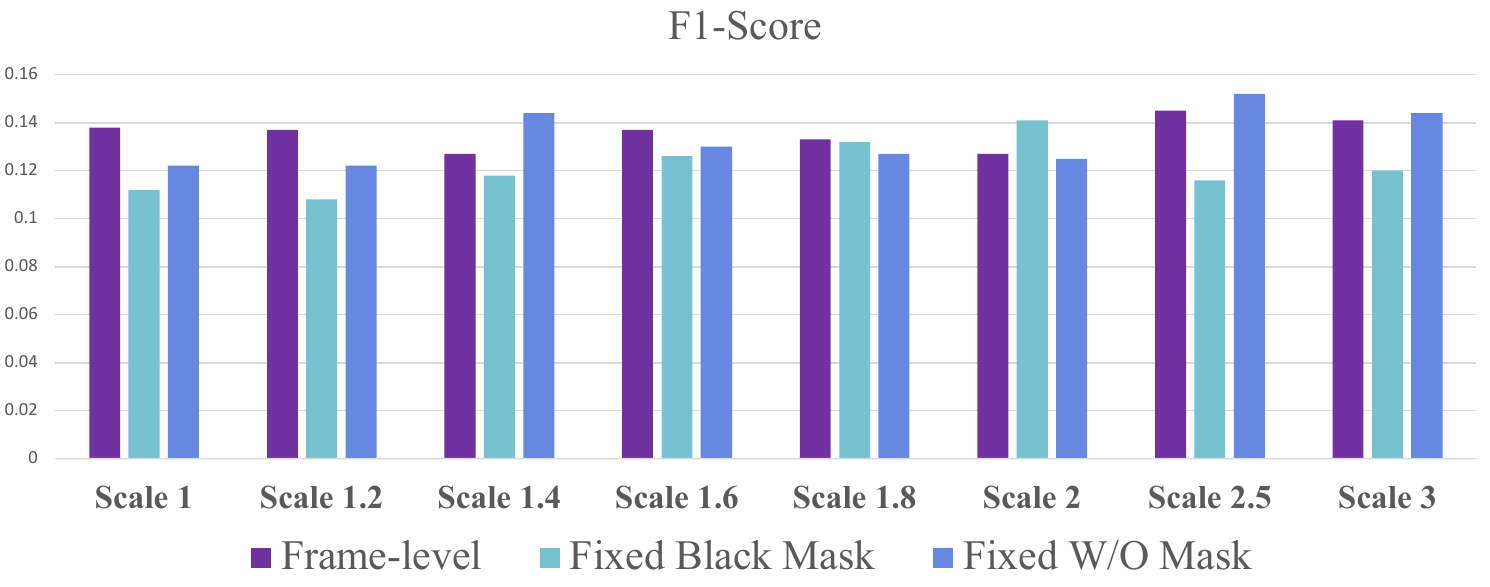}}
\end{center}
\vspace{-1.5em}
  \caption{Exploring diverse cropping scales with MiniGPT-4 at the group level in F1 score.}
\label{fig:scale}
\vspace{-1em}
\end{figure}

\begin{figure*}
\begin{center}
\scalebox{1}{
\includegraphics[width=1\linewidth]{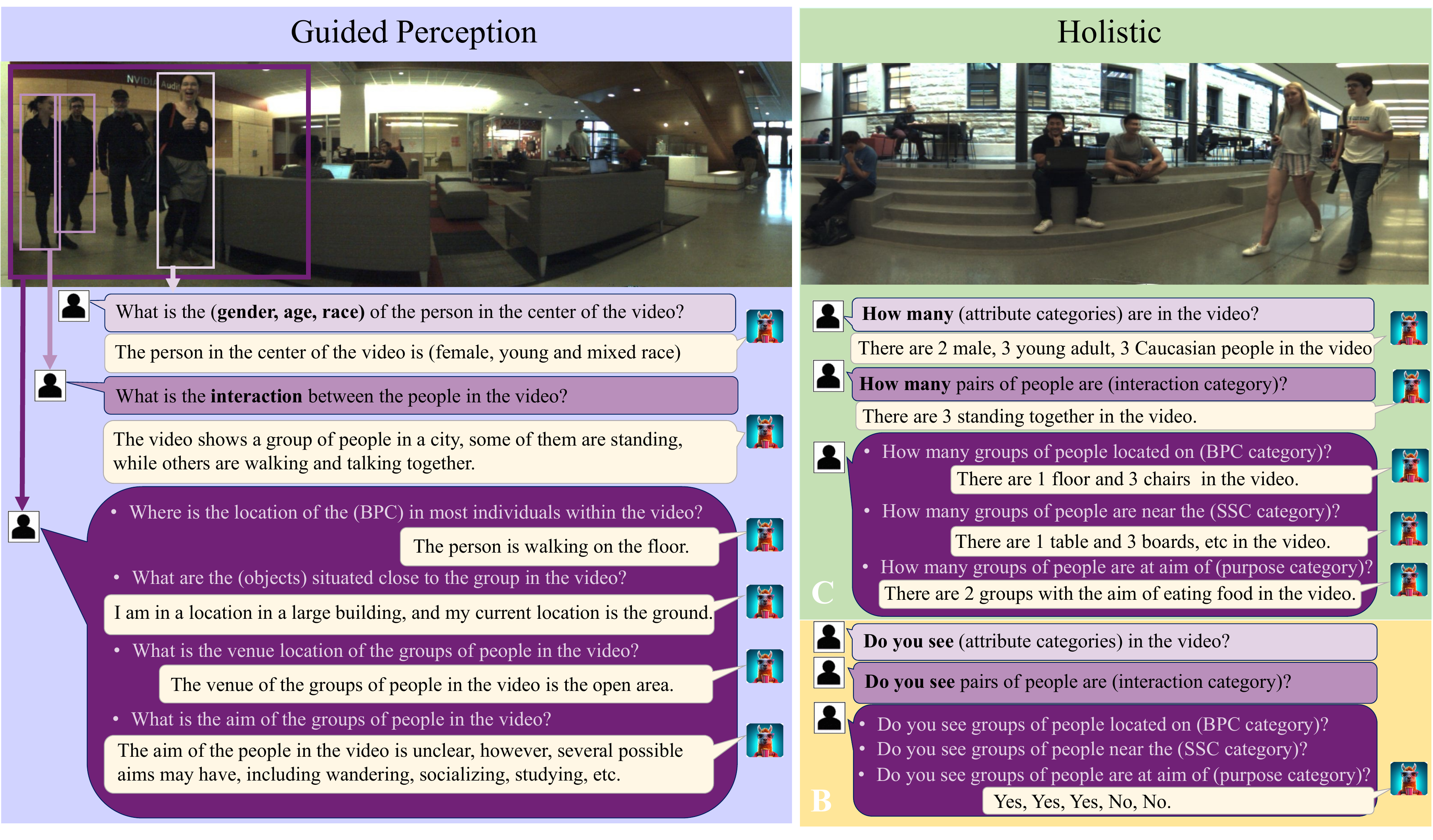}}
\end{center}\vspace{-2em}
  \caption{Illustrating the Guided Perception experiment is depicted through cropped regions delineated by bounding boxes on the left image. The colours—light pink, dark pink, and purple—signify the individual, intra-group, and social group levels, respectively, as detailed in Figure~\ref{fig:tisser}'s colour legend. Holistic experiments are denoted by a green background for the Counting approach (C) and a yellow background for the Binary approach (B).}
\label{fig:prompt}
\vspace{-1em}
\end{figure*}

In this section, we delve into the recent advancements of large language models, particularly their progress in vision-related aspects and multi-modal capabilities. Our objective is to explore the effectiveness of state-of-the-art multi-modal Language Models (LLMs) using the JRDB-Social benchmark. Our focus is to assess their ability to comprehend various complexities of human social behaviour across different difficulty levels and conditions. Specifically, we aim to evaluate their performance at individual, intra-group, and social group levels.
\\
\textbf{Multi-modal LLMs Selection.} 
For our evaluation, we opted for prominent and well-established multi-modal models that have exhibited promising results in recent studies. This selection includes video-based models like Video-LLaMA~\cite{zhang2023video}, VALLEY~\cite{luo2023valley}, and Otter~\cite{li2023otter}. Additionally, our analysis incorporates image-based models, such as MINIGPT-4~\cite{zhu2023minigpt} and InstructBLIP~\cite{instructblip}. This diverse set ensures a comprehensive examination of the current state-of-the-art in multi-modal language models.
\\
\textbf{Metric and Evaluation.}
For evaluating these models based on textual descriptions in JRDB-Social, common metrics like BLEU~\cite{papineni2002bleu}, ROUGE~\cite{chin2004rouge}, and METEOR~\cite{banerjee2005meteor} are often used to measure overall sentence similarity. However, these metrics may lack specificity when the focus is on key entities such as gender, age, aims, venues, etc., embedded in the hard-coded sentence structure. For instance, BLEU and ROUGE lack precision by concentrating on n-gram overlap without considering individual term precision, while METEOR, despite incorporating additional linguistic features, is sensitive to parameter choices.
To sidestep these limitations arising from these metric limitations, we opt to assess the models by prompting questions to extract named entities, such as coloured words in the text description structure, reflecting crucial elements of meaning. We then reformulate the problem as a single or multi-label classification task. This approach aligns with the unique demands of our task, providing a focused and rigorous evaluation framework that addresses the shortcomings of more generic textual metrics.
Also, for evaluating interaction labels, we apply the same metrics. To assess the selected models' performance, we use accuracy and F1 score as metrics. While accuracy measures overall correctness, the F1 score provides a balanced evaluation of precision and recall, particularly valuable in imbalanced scenarios, as observed in the JRDB-Social dataset. While the F1 score results for entire data are outlined here, more comprehensive details and accuracy results are provided in the supplementary materials.\\
\textbf{Experimental Setup and Implementation Details.}
Generally, we conducted two separate experiments, \textbf{Guided Perception} and \textbf{Holistic}, to investigate how multi-modal LLMs perform under different difficulty conditions and levels of guidance.
In the guided perception experiment, we use ground truth bounding boxes to direct the model's focus to specific video regions, providing clear cues for analyzing areas of interest.
In the holistic study, the model is exposed to the entire video without external aids like bounding boxes. This methodology allows the model to conduct a thorough analysis of the video, relying solely on its inherent information, mimicking real-world scenarios where detailed annotations might be lacking.
Figure~\ref{fig:prompt} shows this study on three levels, and more detail is provided in section~\ref{subsec:GuidedPerception} and section~\ref{subsec:Holistic}.\\
To enhance both reliability and performance, we implemented a \textit{Five Ensemble Strategy}. In this strategy, each model undergoes five iterations, and the final output is derived through the utilization of an aggregation strategy. Further details regarding its implementation for both video-based and image-based models can be found in the supplementary materials.
Additionally, in our guided perception experiment for social group analysis, we explored different \textit{cropping scales} to identify the most effective cropping region. Unlike individual or intra-group levels, the model needed to account for a broader context beyond mere bounding boxes. This approach ensured the model's capability to encompass diverse contextual information and maintain robustness across different scenarios, adeptly adapting to scenes featuring both small and large groups. Figure~\ref{fig:scale} displays the various scales using different methods that utilize MiniGPT-4 (model LLaMA-2 7B). Frame-level processing involves cropping videos based on bounding boxes for each frame and resizing them uniformly to \(512 \times 512\) pixels or \(256 \times 256\) pixels. In the fixed black mask method, videos are cropped with non-object areas masked in black. The object's centre point is retained without resizing across frames. The fixed without mask method is akin to the fixed black mask method, but it maintains the full context without using black masking on non-object areas.
Considering the overall F1 average, it was observed that the Frame-level method, with an F1 average of 0.1452 and a scaling factor of 2.5, outperformed both the FixedBlack Mask and Fixed W/O Mask methods. Consequently, the Frame-level method is selected. More details have been provided in the supplementary material.
\subsection{Guided Perception}\label{subsec:GuidedPerception}
In this experiment, we employ ground truth bounding boxes to crop regions of interest. The objective of this approach is to aid the model in localization, directing its attention to specific regions and evaluating its capability to detect the category at three distinct levels. For example, on an individual level, we query the models for the type of gender, age, and race within specific areas of interest for each person. At the intra-group level, we isolate each pair within a group for the entire duration of their interaction. The model's role in this context is to observe the interaction type and discern singular or multiple interactions taking place among these pairs.
At the social group level, the model is presented with each isolated group throughout its entirety. Its task involves recognizing the engagement of body position's connection to the content (BPC), identifying the proximity of significant scene context (SSC), determining the venue where the group is active, and comprehending the group's purpose. These prompts and processes are illustrated in Figure~\ref{fig:prompt}.\\
Based on the results presented in Table~\ref{table:BBf1}, the analysis reveals a consistently reliable performance in predicting individual attributes such as gender and age. However, the detection of race proves to be more intricate, primarily due to the subjective and complex nature of this attribute. Notably, Video-LLaMA and MiniGPT-4 stand out as the top-performing models, attributed to the quality of the data on which they were trained and their design framework. These models exhibit promising results, particularly in tasks related to gender and age prediction. Nevertheless, even these models, experience a decline in performance as the evaluation progresses from the individual to intra-group and social group levels. This observed pattern signifies a significant challenge for the models in comprehending higher-level social contexts. The intricacies associated with attributes like body position’s connection with the content (BPC) and salient scene context (SSC) contribute to the limitations faced by these models, underscoring the ongoing need for advancements to enhance their understanding of diverse and complex social dynamics beyond individual attributes. In this experiment, we explored the model's capabilities by focusing on a limited region. However, the vital question remains: can the model effectively capture intricate details when presented with the entire scene? To answer this query, we delve into a holistic experiment, explained below.
\subsection{Holistic}\label{subsec:Holistic}
In this experiment, the model receives the complete video. The objective is to evaluate how well multi-modal LLMs capture fine-grained information without any cropping or additional assistance and explore their performance across three distinct levels. To this end, we employed two approaches: the counting approach and the binary approach.\\
{\bf Counting Approach.}
In this approach, our central objective is to evaluate the model's ability to identify detailed information and quantify occurrences throughout an entire video. For example, on an individual level, we analyzed the count of females or individuals in young adulthood. At the intra-group level, we inquired about the frequency of diverse interactions between pairs. One instance is the exploration of the number of pairs sitting together in a video, and this investigative process was reiterated for all interaction labels.
Similarly, this methodology is replicated at the social group level. Figure~\ref{fig:prompt} visually depicts this approach.\\
As indicated in Table~\ref{table:Holif1_C}, the experiment demonstrates a significant decline in performance compared to the guided perception experiment, Table~\ref{table:BBf1}. Valley (LLaMA-1 13B) and OTTER (LLaMA-1 7B) models are excluded from Table~\ref{table:BBf1} due to poor performance. This suggests that capturing information at this level of detail poses a substantial challenge for these models. Moreover, the difficulty increases when transitioning to a higher-level social group context, similar to the guided perception experiment. This observation prompted us to simplify the task by assessing whether the model can perceive fine-grained information or not. To explore this, we conducted a binary approach experiment.\\
{\bf Binary Approach.}
As previously mentioned, this approach aims to evaluate the models' ability to capture intricate details without specifying their type and quantity. The purpose of this assessment is to examine their level of understanding achieved by simplifying the task. Similar to the counting approach, the entire video is presented to the model, and the query is simplified to a binary response, either \textit{Yes} or \textit{No}. 
This process is visually depicted in Figure~\ref{fig:prompt}.\ 
Examining the outcomes of this experiment in Table~\ref{table:Holif1_B}, despite not altering the input of the model (entire video), we observed enhanced performance compared to the counting approach, Table~\ref{table:Holif1_C}. The improvement may stem from the hallucination problem~\cite{zhang2023video,li2023otter} present in the text encoder of these models, as the video encoder in this aspect works similarly to the counting approach.
However, even with the simplicity of this approach, the models still encounter challenges in capturing information at the intra-group and social context levels.
This implies that social group contexts pose challenges for multi-modal LLMs, and improvements in various aspects, such as training on more challenging datasets that offer finer-grained information, and developing a more effective framework, are required.

%% file: sec/5_Conclusion.tex
\section{Conclusion}
\Simin{This paper introduces JRDB-Social, a comprehensive robotic dataset designed to investigate human social behaviour within varied contexts. Annotations within the dataset operate across three levels: individual, intra-group, and group, providing detailed attributes, interactions, and contextual descriptions. Leveraging recent advancements in VLMs, the dataset was assessed to gauge their proficiency in understanding human social behaviour in crowded environments. However, findings suggest that VLMs encounter challenges in meaningful visual perception and reasoning on this dataset, particularly in tasks involving complex social interactions. The observed weaknesses may stem from design choices or differences in training data. Thus, there is a need for further advancements in these models to enhance their capability to capture nuanced social understanding within diverse contexts.}
\\ \\
{\bf Acknowledgments.}
The work has received partial funding from The Australian Research Council Discovery Project ARC DP2020102427. Additionally, it is based on research partially sponsored by the DARPA Assured Neuro Symbolic Learning and Reasoning (ANSR) program under award number FA8750-23-2-1016 and the DARPA Computational Cultural Understanding (CCU) program under agreement number HR001122C0029.

%% file: sec_sup/1_dataset.tex
\section{Dataset}
\subsection{Intra-Group Level Dynamic Interactions}
As mentioned in the paper, we presented a protocol to our trained annotators for labeling each interaction. The protocol for annotating each interaction label is outlined below:\\
\textit{Walking Together:} Individuals walking in the same direction with close proximity, either alongside each other or with one person behind the other.
\textit{Walking Toward Each Other:} Individuals walking towards each other, or one person standing while the other approaches.
\textit{Standing Together:} Both individuals standing closely at the same time.
\textit{Moving Together:} One person walking while the other engages in alternative modes such as skating, cycling, or riding a scooter.
\textit{Sitting Together:} Both individuals sitting simultaneously.
\textit{Going Upstairs Together:} Both individuals ascending stairs together at the same time.
\textit{Cycling Together:} Both individuals cycling either alongside each other or in tandem.
\textit{Going Downstairs Together:} Both individuals descending stairs together at the same time.
\textit{Bending Together:} Both individuals bending simultaneously.
\textit{Pointing at Something Together:} Both individuals pointing at something together simultaneously.
\textit{Conversation:} One individual listening while the other talks, or vice versa.
\textit{Looking Into Something Together:} Both individuals looking into something together simultaneously.
\textit{Looking at the Robot:} Both individuals directing their gaze at the robot simultaneously.
\textit{Looking at Something Together:} Both individuals looking at something together at the same time.
\textit{Eating Something Together:} Both individuals consuming food simultaneously.
\textit{Interaction with Door Together:} Both individuals interacting with a door together at the same time.
\textit{Waving Hand:} One or both individuals perform a waving hand gesture, indicating a greeting as specified in individual action labels.
\textit{Shaking Hand:} The individuals shake hands with each other, expressing a greeting as indicated in individual action labels.
\textit{Hugging Each Other:} The individuals embrace each other, conveying a greeting as specified in individual action labels.
\textit{Holding Something Together:} Both individuals holding something together at the same time.
These annotations provide a detailed understanding of various dyadic interactions, offering valuable insights.
\subsection{Social Group Level Context}
\textbf{Engagement of Body Position with the Content and Salient Scene Content.} In this section, we discussed the association between body positions and content (BPC) and the identification of salient scene content in proximity to a group (SSC). The category of annotations for both BPC and SSC are detailed below:\\
\textbf{BPC Annotations:}
\textit{floor}, \textit{ground}, \textit{chair}, \textit{sidewalk}, \textit{bike}, \textit{stairs}, \textit{platform}, \textit{sofa}, \textit{grass}, \textit{street}, \textit{crosswalk}, \textit{road}, \textit{scooter}, \textit{skateboard}, \textit{pathway}, \textit{desk}, \textit{balcony}, \textit{bench}.\\
\textbf{SSC Annotations:}
\textit{gate}, \textit{table}, \textit{counter}, \textit{door}, \textit{pillar}, \textit{shelves}, \textit{wall}, \textit{standboard}, \textit{poster}, \textit{desk}, \textit{food-truck}, \textit{bike}, \textit{chair}, \textit{stairs}, \textit{fence}, \textit{show-case}, \textit{room}, \textit{board}, \textit{cabinet}, \textit{garbage-bin}, \textit{stroller}, \textit{elevator}, \textit{buffet-cafeteria}, \textit{trolley}, \textit{forecourt}, \textit{scooter}, \textit{bus}, \textit{robot}, \textit{platform}, \textit{window}, \textit{tree}, \textit{pole}, \textit{crutches}, \textit{stand-pillar}, \textit{screen}, \textit{car}, \textit{copy-machine}, \textit{class}, \textit{coffee-machine}, \textit{balcony}, \textit{sofa}, \textit{statue}, \textit{floor}, \textit{bench}, \textit{building}, \textit{baggage}, \textit{shop}, \textit{light-street}, \textit{drink-fountain}.

%% file: sec_sup/2_experiments.tex
\begin{table}[b]
    \centering
    \scalebox{0.9}{
\begin{tabular}{cc|cccc}
    \toprule
    Experiments & Metrics & Train & Validation & Test \\ \hline
    \multirow{2}{*}{Guided Perception} 
    & F1 score & \ref{table:GP-F1-train} & \ref{table:GP-F1-valid} & \ref{table:GP-F1-test} \\
    & Accuracy & \ref{table:GPTr} & \ref{table:GPV} & \ref{table:GPTe} \\ \hline
    \multirow{2}{*}{Holistic (Counting)} & F1 score & \ref{table:HCTr-F1} & \ref{table:HCV-F1} & \ref{table:HCTe-F1} \\
    & Accuracy & \ref{table:HCTr} & \ref{table:HCV} & \ref{table:HCTe} \\ \hline
    \multirow{2}{*}{Holistic (Binary)} & F1 score & \ref{table:HBTr-F1-score} & \ref{table:HBV-F1-score} & \ref{table:HBTe-F1} \\
    & Accuracy & \ref{table:HBTr} & \ref{table:HBV} & \ref{table:HBTe} \\\hline
    \end{tabular}}
    \caption{Summary of Table Contents.}
    \label{tab:content_of_tables}
\end{table}

\input{sec_sup/X_tables_figures}

\input{sec_sup/Y_prompts}

\section{Experiments}
In the paper, we presented F1 scores for the entire dataset. In this section, we provide F1 scores in Table~\ref{table:GP-F1-train}, \ref{table:GP-F1-valid}, \ref{table:GP-F1-test}, \ref{table:HCTr-F1}, \ref{table:HCV-F1}, \ref{table:HCTe-F1}, \ref{table:HBTr-F1-score}, \ref{table:HBV-F1-score}, \ref{table:HBTe-F1}, and accuracy in Table~\ref{table:GPTr}, \ref{table:GPV}, \ref{table:GPTe}, \ref{table:HCTr}, \ref{table:HCV}, \ref{table:HCTe}, \ref{table:HBTr}, \ref{table:HBV} and \ref{table:HBTe}. These metrics are reported separately for the training, validation, and test datasets for both Guided Perception and Holistic approaches. The summaries of all tables are provided in Table \ref{tab:content_of_tables}. Furthermore, as previously stated, we implemented a Five Ensemble Strategy, elaborated upon in Table~\ref{table:scale} for both F1 Score and accuracy.
\vspace{-10em}

%% file: sec_sup/X_tables_figures.tex
\begin{table*}[b]
   \centering
   \footnotesize
   \scalebox{0.9}{
   \begin{tabular}{lccc|c|cccc}
    \toprule
    {\bf Multi-modal LLM} & \multicolumn{3}{c}{ {\bf Individual Level}} & \multicolumn{1}{c}{ {\bf Intra-Group Level }} & \multicolumn{4}{c}{{\bf Social Group Level}} \\
    \cline{2-9}
    & Gender & Age & Race & Interactions & BPC & SSC & Venue & Purpose \\
    \hline
    Video-LLaMA (LLaMA-2 13B)~\cite{zhang2023video} (5 Ens)& 0.7904 & 0.3414 & 0.2970 & 0.3439 & 0.1595 &0.1176& 0.1430 & 0.2785   \\
    Video-LLaMA (LLaMA-2 13B)~\cite{zhang2023video}  & 0.7153 & 0.2260 & 0.2587 & 0.3129 & 0.0978 &0.0875  & 0.1305 & 0.2651   \\
    Video-LLaMA (LLaMA-2 7B)~\cite{zhang2023video} (5 Ens) &  0.6086& 0.3999 &  0.2551& 0.4261 & 0.1668 & 0.1176 &0.1691  & 0.3039   \\
    Video-LLaMA (LLaMA-2 7B)~\cite{zhang2023video} &  0.5754& 0.2961 & 0.2407 & 0.3944 & 0.0847 &0.0866  & 0.1593 & 0.2918  \\
    Valley (LLaMA-1 13B)~\cite{luo2023valley} (5 Ens) &  0.5321&0.1957&0.1323&0.1620& 0.0945 &0.0308 & 0.0360 & 0.2205  \\
    Valley (LLaMA-1 13B)~\cite{luo2023valley} & 0.5426 & 0.1944&  0.0752&0.1884  & 0.0224 & 0.0449 &  0.0418& 0.2510   \\
\hline
  \end{tabular}
   }
\caption{{\bf Guided Perception}: Comparing popular multi-modal LLMs across the JRDB-Social at three levels in {\bf F1-score} for the {\bf train set}. BPC = Engagement of Body Position’s connection with the Content, SSC = Salient Scene Content, (5 Ens) = Five Ensemble Strategy.}

\label{table:GP-F1-train}
\end{table*}
\begin{table*}[b]
   \centering
   \footnotesize
   \scalebox{0.9}{
   \begin{tabular}{lccc|c|cccc}
    \toprule
    {\bf Multi-modal LLM} & \multicolumn{3}{c}{ {\bf Individual Level}} & \multicolumn{1}{c}{ {\bf Intra-Group Level }} & \multicolumn{4}{c}{{\bf Social Group Level}} \\
    \cline{2-9}
    & Gender & Age & Race & Interactions & BPC & SSC & Venue & Purpose \\
    \hline
    Video-LLaMA (LLaMA-2 13B)~\cite{zhang2023video} (5 Ens) & 0.6960 & 0.3201 &0.5253& 0.3322& 0.1200 & 0.2153 & 0.2379 & 0.1968 \\
    Video-LLaMA (LLaMA-2 13B)~\cite{zhang2023video} &0.6872 & 0.1996 & 0.2827 & 0.3368 & 0.0868 & 0.1181 & 0.2200 & 0.2128 \\
    Video-LLaMA (LLaMA-2 7B)~\cite{zhang2023video} (5 Ens) & 0.5189& 0.3888 & 0.3745 & 0.3434 & 0.1420 &0.1878 & 0.1702 &0.2634 \\
    Video-LLaMA (LLaMA-2 7B)~\cite{zhang2023video} & 0.5051 &0.2279 &0.2348&0.3781  &0.1182 &0.1138& 0.2018 & 0.2336 \\
    Valley (LLaMA-1 13B)~\cite{luo2023valley} (5 Ens) &0.5492  &0.2341&  0.1555& 0.1302 & 0.0279 & 0.0369 & 0.1409& 0.2548 \\
    Valley (LLaMA-1 13B)~\cite{luo2023valley} &0.5199 &0.2313& 0.0812&0.1519& 0.0110 &0.0325& 0.1244 &0.2348 \\
\hline
  \end{tabular}}
\caption{{\bf Guided Perception}: Comparing popular multi-modal LLMs across the JRDB-Social at three levels in {\bf F1-Score}for the {\bf validation set}. BPC = Engagement of Body Position’s connection with the Content, SSC = Salient Scene Content, (5 Ens) = Five Ensemble Strategy.}
\label{table:GP-F1-valid}
\end{table*}
\begin{table*}[b]
   \centering
   \footnotesize
   \scalebox{0.9}{
   \begin{tabular}{lccc|c|cccc}
    \toprule
    {\bf Multi-modal LLM} & \multicolumn{3}{c}{ {\bf Individual Level}} & \multicolumn{1}{c}{ {\bf Intra-Group Level }} & \multicolumn{4}{c}{{\bf Social Group Level}} \\
    \cline{2-9}
    & Gender & Age & Race & Interactions & BPC & SSC & Venue & Purpose \\
    \hline
    Video-LLaMA (LLaMA-2 13B)~\cite{zhang2023video} (5 Ens) & 0.6626 & 0.2768 & 0.2275 & 0.2868 & 0.1898 & 0.0932 & 0.3109 & 0.2628 \\
    Video-LLaMA (LLaMA-2 13B)~\cite{zhang2023video} & 0.6465& 0.2065 & 0.1947 & 0.2584& 0.1166 & 0.1696 & 0.2764 & 0.2400\\
    Video-LLaMA (LLaMA-2 7B)~\cite{zhang2023video} (5 Ens) & 0.4554 & 0.2385& 0.1854 & 0.3319 &0.1701 & 0.2455 & 0.2579 &0.3073 \\
    Video-LLaMA (LLaMA-2 7B)~\cite{zhang2023video} & 0.4601 & 0.1915 & 0.1837 & 0.3153 &0.1555 & 0.1551 & 0.2973 & 0.2762 \\
    Valley (LLaMA-1 13B)~\cite{luo2023valley} (5 Ens) & 0.2617 & 0.2022 & 0.1143 & 0.1435 & 0.0503 & 0.0385 & 0.1155 & 0.2694 \\
    Valley (LLaMA-1 13B)~\cite{luo2023valley} & 0.2862 &0.1940 & 0.0752 & 0.1606& 0.0354 & 0.0480 & 0.1395 & 0.2520\\
\hline
  \end{tabular}}
\caption{ {\bf Guided Perception}: Comparing popular multi-modal LLMs across the JRDB-Social at three levels in {\bf F1-score} for the {\bf test set}. BPC = Engagement of Body Position’s connection with the Content, SSC = Salient Scene Content, (5 Ens) = Five Ensemble Strategy.
}
\label{table:GP-F1-test}
\end{table*}
\begin{table*}[b]
   \centering
   \footnotesize
   \scalebox{0.9}{
   \begin{tabular}{lccc|c|cccc}
    \toprule
    {\bf Multi-modal LLM} & \multicolumn{3}{c}{ {\bf Individual Level}} & \multicolumn{1}{c}{ {\bf Intra-Group Level }} & \multicolumn{4}{c}{{\bf Social Group Level}} \\
    \cline{2-9}
    & Gender & Age & Race & Interactions & BPC & SSC & Venue & Purpose \\
    \hline
    Video-LLaMA (LLaMA-2 13B)~\cite{zhang2023video} (5 Ens)& 0.3486 & 0.2048 & 0.3454 & 0.0906 & 0.0681 &0.0321 & 0.2572 & 0.1520 \\
    Video-LLaMA (LLaMA-2 13B)~\cite{zhang2023video}  & 0.1601 & 0.1999 & 0.2699 & 0.0692 &0.0797 &0.0338 & 0.3070 & 0.1247 \\
    Video-LLaMA (LLaMA-2 7B)~\cite{zhang2023video} (5 Ens) & 0.2378 &0.1809& 0.3032 & 0.0854 &0.0566 &0.0269 & 0.2303 & 0.1224 \\
    Video-LLaMA (LLaMA-2 7B)~\cite{zhang2023video} & 0.0704 &0.0815 & 0.1706 & 0.0359 & 0.0354 &0.0208 & 0.2187 & 0.0360 \\
    Valley (LLaMA-1 13B)~\cite{luo2023valley} (5 Ens) & 0.0006 & 0.0094& 0.0225 & 0.0009 & 0.0000 & 0.0000 & 0.1315 & 0.0007 \\
    Valley (LLaMA-1 13B)~\cite{luo2023valley} & 0.0000 & 0.0090 & 0.0090 & 0.0020 & 0.0000 & 0.0000 &0.1362 & 0.0000 \\
\hline
  \end{tabular}}
\caption{{\bf Holistic (Counting) Experiment}: Comparing popular multi-modal LLMs across the JRDB-Social at three levels in {\bf F1-Score} for the {\bf train set}. BPC = Engagement of Body Position’s connection with the Content, SSC = Salient Scene Content, (5 Ens) = Five Ensemble Strategy.}
\label{table:HCTr-F1}
\end{table*}
\begin{table*}[b]
   \centering
   \footnotesize
   \scalebox{0.9}{
   \begin{tabular}{lccc|c|cccc}
    \toprule
    {\bf Multi-modal LLM} & \multicolumn{3}{c}{ {\bf Individual Level}} & \multicolumn{1}{c}{ {\bf Intra-Group Level }} & \multicolumn{4}{c}{{\bf Social Group Level}} \\
    \cline{2-9}
    & Gender & Age & Race & Interactions & BPC & SSC & Venue & Purpose \\
    \hline
    Video-LLaMA (LLaMA-2 13B)~\cite{zhang2023video} (5 Ens)& 0.3893 & 0.2759 & 0.3335 & 0.0866 & 0.0955 & 0.0437 & 0.2500 & 0.1722 \\
    Video-LLaMA (LLaMA-2 13B)~\cite{zhang2023video}  & 0.1553 & 0.2820 & 0.2488 & 0.0559 & 0.0992 & 0.0508 &0.2448 & 0.1189 \\
    Video-LLaMA (LLaMA-2 7B)~\cite{zhang2023video} (5 Ens)  & 0.1927 & 0.2384 & 0.2991 &0.0563 &0.1026 & 0.0376 & 0.1295 &0.1597 \\
    Video-LLaMA (LLaMA-2 7B)~\cite{zhang2023video} & 0.0356 & 0.1353 & 0.1424 & 0.0174 & 0.0584 & 0.0268 & 0.1922 & 0.0592 \\
    Valley (LLaMA-1 13B)~\cite{luo2023valley} (5 Ens) & 0.0000 &0.0303 & 0.0241 & 0.0055 & 0.0000 & 0.0000 & 0.1547 & 0.0000 \\
    Valley (LLaMA-1 13B)~\cite{luo2023valley} & 0.0000 & 0.0161 & 0.0002 & 0.0000 & 0.0000 & 0.0000 & 0.0630 & 0.0000 \\
\hline
  \end{tabular}}
\caption{ {\bf Holistic (Counting) Experiment}: Comparing popular multi-modal LLMs across the JRDB-Social at three levels in {\bf F1-Score} for the {\bf validation set}. BPC = Engagement of Body Position’s connection with the Content, SSC = Salient Scene Content, (5 Ens) = Five Ensemble Strategy.}
\label{table:HCV-F1}
\end{table*}
\begin{table*}[b]
   \centering
   \footnotesize
   \scalebox{0.9}{
   \begin{tabular}{lccc|c|cccc}
    \toprule
    {\bf Multi-modal LLM} & \multicolumn{3}{c}{ {\bf Individual Level}} & \multicolumn{1}{c}{ {\bf Intra-Group Level }} & \multicolumn{4}{c}{{\bf Social Group Level}} \\
    \cline{2-9}
    & Gender & Age & Race & Interactions & BPC & SSC & Venue & Purpose \\
    \hline
    Video-LLaMA (LLaMA-2 13B)~\cite{zhang2023video} (5 Ens)& 0.3086 & 0.2854 & 0.3591 & 0.1164 & 0.0838 & 0.0125 & 0.3150 & 0.2011 \\
    Video-LLaMA (LLaMA-2 13B)~\cite{zhang2023video}  & 0.1540 & 0.2636 & 0.2702 & 0.0803 & 0.1025 & 0.0146 & 0.2913 & 0.1414 \\
    Video-LLaMA (LLaMA-2 7B)~\cite{zhang2023video} (5 Ens) & 0.2093 & 0.2555 & 0.2947 & 0.0878 & 0.0588 & 0.0094 & 0.3804 & 0.1430 \\
    Video-LLaMA (LLaMA-2 7B)~\cite{zhang2023video} & 0.0000 & 0.0095 & 0.0048 & 0.0046 & 0.0000 & 0.0000 & 0.2148 & 0.0094 \\
    Valley (LLaMA-1 13B)~\cite{luo2023valley} (5 Ens) &0.0017 & 0.0303 & 0.0263 & 0.0000 & 0.0000 & 0.0000 & 0.0863 & 0.0010 \\
    Valley (LLaMA-1 13B)~\cite{luo2023valley} & 0.0017 & 0.0104 & 0.0043 & 0.0000 & 0.0015 & 0.0000 & 0.1037 & 0.0000 \\
\hline
  \end{tabular}}
\caption{{\bf Holistic (Counting) Experiment}: Comparing popular multi-modal LLMs across the JRDB-Social at three levels in {\bf F1-Score} for the {\bf test set}. BPC = Engagement of Body Position’s connection with the Content, SSC = Salient Scene Content, (5 Ens) = Five Ensemble Strategy.}
\label{table:HCTe-F1}
\end{table*}
\begin{table*}[b]
   \centering
   \footnotesize
   \scalebox{0.9}{
   \begin{tabular}{lccc|c|cccc}
    \toprule
    {\bf Multi-modal LLM} & \multicolumn{3}{c}{ {\bf Individual Level}} & \multicolumn{1}{c}{ {\bf Intra-Group Level }} & \multicolumn{4}{c}{{\bf Social Group Level}} \\
    \cline{2-9}
    & Gender & Age & Race & Interactions & BPC & SSC & Venue & Purpose \\
    \hline
    Video-LLaMA (LLaMA-2 13B)~\cite{zhang2023video} (5 Ens)& 0.9664 & 0.4984 & 0.7341 & 0.3013 & 0.2066 & 0.0976 & 0.2733& 0.4272 \\
    Video-LLaMA (LLaMA-2 13B)~\cite{zhang2023video}  & 0.9637 &0.4992& 0.7333 & 0.3112 &0.2136& 0.1017 &0.2733&0.4207\\
    Video-LLaMA (LLaMA-2 7B)~\cite{zhang2023video} (5 Ens) & 0.9664 & 0.4984 &0.7341& 0.2920 & 0.2000 &0.0959 & 0.2733 & 0.4092 \\
    Video-LLaMA (LLaMA-2 7B)~\cite{zhang2023video} &0.9664  &0.4984 & 0.7341 & 0.2942 & 0.2003 &0.0960& 0.2733 & 0.4119 \\
    Valley (LLaMA-1 13B)~\cite{luo2023valley} (5 Ens) & 0.9243 & 0.5048 &0.7305 & 0.3238 & 0.2398 & 0.1107 & 0.2723 & 0.3260 \\
    Valley (LLaMA-1 13B)~\cite{luo2023valley} & 0.8685& 0.5224 &0.7293  &  0.3048&0.1958&0.1123&0.2647&0.3740 \\
\hline
  \end{tabular}}
\caption{ {\bf Holistic (Binary) Experiment}: Comparing popular multi-modal LLMs across the JRDB-Social at three levels in {\bf F1-Score} for the {\bf train set}. BPC = Engagement of Body Position’s connection with the Content, SSC = Salient Scene Content, (5 Ens) = Five Ensemble Strategy.}
\label{table:HBTr-F1-score}
\end{table*}
\begin{table*}[b]
   \centering
   \footnotesize
   \scalebox{0.9}{
   \begin{tabular}{lccc|c|cccc}
    \toprule
    {\bf Multi-modal LLM} & \multicolumn{3}{c}{ {\bf Individual Level}} & \multicolumn{1}{c}{ {\bf Intra-Group Level }} & \multicolumn{4}{c}{{\bf Social Group Level}} \\
    \cline{2-9}
    & Gender & Age & Race & Interactions & BPC & SSC & Venue & Purpose \\
    \hline
    Video-LLaMA (LLaMA-2 13B)~\cite{zhang2023video} (5 Ens)& 0.9855 & 0.5477 & 0.7155 & 0.3183  & 0.2848 & 0.1236 & 0.2857 & 0.4774 \\
    Video-LLaMA (LLaMA-2 13B)~\cite{zhang2023video}  & 0.9855 & 0.5523 & 0.7069 & 0.3097 &  0.2879 & 0.1266 &0.2857 & 0.5149 \\
    Video-LLaMA (LLaMA-2 7B)~\cite{zhang2023video} (5 Ens) & 0.9855& 0.5477 & 0.7155 & 0.3125 & 0.2739 & 0.1174 & 0.2857 &0.4792 \\
    Video-LLaMA (LLaMA-2 7B)~\cite{zhang2023video} & 0.9855 & 0.5477 & 0.7155 & 0.3127 & 0.2739 & 0.1175 & 0.2857 & 0.4801 \\
    Valley (LLaMA-1 13B)~\cite{luo2023valley} (5 Ens) &0.9393&0.5084 & 0.7180 &  0.3125& 0.2727 & 0.1295 & 0.2880 & 0.4435 \\
    Valley (LLaMA-1 13B)~\cite{luo2023valley} & 0.8333 &0.4691 & 0.7188 & 0.3177 & 0.2379 &0.1339&  0.2881& 0.4189\\
\hline
  \end{tabular}}
\caption{ {\bf Holistic (Binary) Experiment}: Comparing popular multi-modal LLMs across the JRDB-Social at three levels in {\bf F1-Score} for the {\bf validation set}. BPC = Engagement of Body Position’s connection with the Content, SSC = Salient Scene Content, (5 Ens) = Five Ensemble Strategy.}
\label{table:HBV-F1-score}
\end{table*}
\begin{table*}[b]
   \centering
   \footnotesize
   \scalebox{0.9}{
   \begin{tabular}{lccc|c|cccc}
    \toprule
    {\bf Multi-modal LLM} & \multicolumn{3}{c}{ {\bf Individual Level}} & \multicolumn{1}{c}{ {\bf Intra-Group Level }} & \multicolumn{4}{c}{{\bf Social Group Level}} \\
    \cline{2-9}
    & Gender & Age & Race & Interactions & BPC & SSC & Venue & Purpose \\
    \hline
    Video-LLaMA (LLaMA-2 13B)~\cite{zhang2023video} (5 Ens)& 0.9906 & 0.6160 & 0.7643 & 0.3542 & 0.2223 & 0.0450 & 0.2857 & 0.5172 \\
    Video-LLaMA (LLaMA-2 13B)~\cite{zhang2023video}  & 0.9906 &0.6184 & 0.7651 & 0.3610  & 0.2324 & 0.0461 & 0.2863 &0.5125 \\
    Video-LLaMA (LLaMA-2 7B)~\cite{zhang2023video} (5 Ens) & 0.9906 & 0.6153 &0.7671 & 0.3394 & 0.2158 &0.0446 & 0.2857& 0.5007 \\
    Video-LLaMA (LLaMA-2 7B)~\cite{zhang2023video} & 0.9887 & 0.6153 & 0.7671 & 0.3436 & 0.2160 & 0.0443 & 0.2857 & 0.5030 \\
    Valley (LLaMA-1 13B)~\cite{luo2023valley} (5 Ens) & 0.9304 & 0.5342 & 0.7688 & 0.3701 & 0.2077 & 0.0493 & 0.2893 & 0.3830 \\
    Valley (LLaMA-1 13B)~\cite{luo2023valley} & 0.8884 & 0.5233 & 0.7647 & 0.3525 &0.2073 & 0.0487 & 0.2838 & 0.4004 \\
\hline
  \end{tabular}}
\caption{ {\bf Holistic (Binary) Experiment}: Comparing popular multi-modal LLMs across the JRDB-Social at three levels in {\bf F1-Score} for the {\bf test set}. BPC = Engagement of Body Position’s connection with the Content, SSC = Salient Scene Content, (5 Ens) = Five Ensemble Strategy.}
\label{table:HBTe-F1}
\end{table*}

\subsection{Prompting}
In the Experiment section of the paper, we initially presented the prompt schematic. 
Prompts~\hyperref[prompt:promptGP]{1}, \hyperref[prompt:promptHC]{2}, and \hyperref[prompt:promptHB]{3} illustrate the guided perception experiment, holistic experiment (counting approach), and holistic experiment (binary approach), respectively.
\begin{table*}[b]
   \centering
   \footnotesize
   \scalebox{0.9}{
\begin{tabular}{lccc|c|cccc}
    \toprule
    {\bf Multi-modal LLM} & \multicolumn{3}{c}{ {\bf Individual Level}} & \multicolumn{1}{c}{ {\bf Intra-Group Level }} & \multicolumn{4}{c}{{\bf Social Group Level}} \\
    \cline{2-9}
    & Gender & Age & Race & Interactions & BPC & SSC & Venue & Purpose \\
    \hline
                            Video-LLaMA (LLaMA-2 13B)~\cite{zhang2023video} (5 Ens)& 0.8675 & 0.8874 & 0.5661 & 0.8495 & 0.7128 &0.9010& 0.1965 & 0.7741   \\
    Video-LLaMA (LLaMA-2 13B)~\cite{zhang2023video}  & 0.8188 & 0.5916 & 0.4261 & 0.8505 & 0.4940 &0.8359  & 0.1780 & 0.7532   \\
    Video-LLaMA (LLaMA-2 7B)~\cite{zhang2023video} (5 Ens) &  0.7909& 0.9489 &  0.5461& 0.8901 & 0.5434 & 0.5434 &0.1719  & 0.8528   \\
    Video-LLaMA (LLaMA-2 7B)~\cite{zhang2023video} &  0.7549& 0.7703 & 0.4815 & 0.8804 & 0.4910 &0.7855  & 0.2088 & 0.8159  \\
    Valley (LLaMA-1 13B)~\cite{luo2023valley} (5 Ens) &  0.7630&0.9535&0.1984&0.5376& 0.1769 &0.9628 & 0.0644 & 0.8336   \\
    Valley (LLaMA-1 13B)~\cite{luo2023valley} & 0.7630 & 0.9396&  0.1015&0.6261  & 0.0614 & 0.9414 &  0.0818& 0.8229   \\
\hline
  \end{tabular}
   }
\caption{{\bf Guided Perception}: Comparing popular multi-modal LLMs across the JRDB-Social at three levels in {\bf accuracy} for the {\bf train set}. BPC = Engagement of Body Position’s connection with the Content, SSC = Salient Scene Content, (5 Ens) = Five Ensemble Strategy.}

\label{table:GPTr}
\end{table*}
\begin{table*}[b]
   \centering
   \footnotesize
   \scalebox{0.9}{
   \begin{tabular}{lccc|c|cccc}
    \toprule
    {\bf Multi-modal LLM} & \multicolumn{3}{c}{ {\bf Individual Level}} & \multicolumn{1}{c}{ {\bf Intra-Group Level }} & \multicolumn{4}{c}{{\bf Social Group Level}} \\
    \cline{2-9}
    & Gender & Age & Race & Interactions & BPC & SSC & Venue & Purpose \\
    \hline
    Video-LLaMA (LLaMA-2 13B)~\cite{zhang2023video} (5 Ens) & 0.7795 & 0.7603 &0.6220& 0.6236 & 0.5707 & 0.9288 & 0.5630 & 0.7661 \\
    Video-LLaMA (LLaMA-2 13B)~\cite{zhang2023video} & 0.7731 & 0.5047 & 0.4418 & 0.6729 & 0.3863 & 0.8467 & 0.4761 & 0.7427 \\
    Video-LLaMA (LLaMA-2 7B)~\cite{zhang2023video} (5 Ens) & 0.7150& 0.8498 & 0.5465 & 0.8610 & 0.3636 &0.8936 & 0.5154 &0.8479 \\
    Video-LLaMA (LLaMA-2 7B)~\cite{zhang2023video} & 0.6932 &0.7188 &0.5058&0.7297  &0.3888 &0.8164& 0.5126 & 0.8082 \\
    Valley (LLaMA-1 13B)~\cite{luo2023valley} (5 Ens) &0.7667  &0.5271&  0.5523& 0.8819 & 0.5732 & 0.9653 &  0.4817& 0.1102 \\
    Valley (LLaMA-1 13B)~\cite{luo2023valley} & 0.6932 &0.8498& 0.1046 &0.5450& 0.0277 &0.9020& 0.4425 &0.8304 \\
\hline
  \end{tabular}}
\caption{{\bf Guided Perception}: Comparing popular multi-modal LLMs across the JRDB-Social at three levels in {\bf accuracy} for the {\bf validation set}. BPC = Engagement of Body Position’s connection with the Content, SSC = Salient Scene Content, (5 Ens) = Five Ensemble Strategy.}
\label{table:GPV}
\end{table*}
\begin{table*}[b]
   \centering
   \footnotesize
   \scalebox{0.9}{
   \begin{tabular}{lccc|c|cccc}
    \toprule
    {\bf Multi-modal LLM} & \multicolumn{3}{c}{ {\bf Individual Level}} & \multicolumn{1}{c}{ {\bf Intra-Group Level }} & \multicolumn{4}{c}{{\bf Social Group Level}} \\
    \cline{2-9}
    & Gender & Age & Race & Interactions & BPC & SSC & Venue & Purpose \\
    \hline
    Video-LLaMA (LLaMA-2 13B)~\cite{zhang2023video} (5 Ens) & 0.7609 & 0.6896 & 0.4215 & 0.8361 & 0.5094 & 0.6199 & 0.5619 & 0.7795 \\
    Video-LLaMA (LLaMA-2 13B)~\cite{zhang2023video} & 0.7413 & 0.4674 & 0.3296 & 0.8425 & 0.3541 & 0.8731 & 0.5004 & 0.7520 \\
    Video-LLaMA (LLaMA-2 7B)~\cite{zhang2023video} (5 Ens) & 0.6723 & 0.7154 & 0.4035 & 0.8597 & 0.4495 & 0.9048 & 0.4035 & 0.8524 \\
    Video-LLaMA (LLaMA-2 7B)~\cite{zhang2023video} & 0.6485 & 0.5566 & 0.3475 & 0.8557 & 0.3996 & 0.8349 & 0.4796 & 0.8095 \\
    Valley (LLaMA-1 13B)~\cite{luo2023valley} (5 Ens) & 0.6110 & 0.7288 & 0.1535 & 0.4947 & 0.1165 & 0.9700 & 0.4001 & 0.8441 \\
    Valley (LLaMA-1 13B)~\cite{luo2023valley} & 0.6306 & 0.7596 & 0.0862 & 0.5913 & 0.0688 & 0.9446 & 0.3972 & 0.8263 \\
\hline
  \end{tabular}}
\caption{ {\bf Guided Perception}: Comparing popular multi-modal LLMs across the JRDB-Social at three levels in {\bf accuracy} for the {\bf test set}. BPC = Engagement of Body Position’s connection with the Content, SSC = Salient Scene Content, (5 Ens) = Five Ensemble Strategy.
}
\label{table:GPTe}
\end{table*}
\begin{table*}[b]
   \centering
   \footnotesize
   \scalebox{0.9}{
   \begin{tabular}{lccc|c|cccc}
    \toprule
    {\bf Multi-modal LLM} & \multicolumn{3}{c}{ {\bf Individual Level}} & \multicolumn{1}{c}{ {\bf Intra-Group Level }} & \multicolumn{4}{c}{{\bf Social Group Level}} \\
    \cline{2-9}
    & Gender & Age & Race & Interactions & BPC & SSC & Venue & Purpose \\
    \hline
    Video-LLaMA (LLaMA-2 13B)~\cite{zhang2023video} (5 Ens)& 0.2634 & 0.1575 & 0.2780 & 0.0711 & 0.0531 & 0.0257 & 0.3158 & 0.1173 \\
    Video-LLaMA (LLaMA-2 13B)~\cite{zhang2023video}  & 0.1196 & 0.1556 & 0.2151 & 0.0517 & 0.0603 & 0.0270 & 0.3579 & 0.0969 \\
    Video-LLaMA (LLaMA-2 7B)~\cite{zhang2023video} (5 Ens) & 0.1799 & 0.1380 & 0.2404 & 0.0662 & 0.0444 & 0.0212 & 0.2632 & 0.0943 \\
    Video-LLaMA (LLaMA-2 7B)~\cite{zhang2023video} & 0.0545 & 0.0612 & 0.1344 & 0.0280 & 0.0259 & 0.0166 & 0.2632 & 0.0283 \\
    Valley (LLaMA-1 13B)~\cite{luo2023valley} (5 Ens) & 0.0004 & 0.0067 & 0.0177 & 0.0017 & 0.0000 & 0.0000 & 0.1790 & 0.0004 \\
    Valley (LLaMA-1 13B)~\cite{luo2023valley} & 0.0000 & 0.0060 & 0.0066 & 0.0017 & 0.0000 & 0.0000 & 0.1790 & 0.0000 \\
\hline
  \end{tabular}}
\caption{{\bf Holistic (Counting) Experiment}: Comparing popular multi-modal LLMs across the JRDB-Social at three levels in {\bf accuracy} for the {\bf train set}. BPC = Engagement of Body Position’s connection with the Content, SSC = Salient Scene Content, (5 Ens) = Five Ensemble Strategy.}
\label{table:HCTr}
\end{table*}
\begin{table*}[b]
   \centering
   \footnotesize
   \scalebox{0.9}{
   \begin{tabular}{lccc|c|cccc}
    \toprule
    {\bf Multi-modal LLM} & \multicolumn{3}{c}{ {\bf Individual Level}} & \multicolumn{1}{c}{ {\bf Intra-Group Level }} & \multicolumn{4}{c}{{\bf Social Group Level}} \\
    \cline{2-9}
    & Gender & Age & Race & Interactions & BPC & SSC & Venue & Purpose \\
    \hline
    Video-LLaMA (LLaMA-2 13B)~\cite{zhang2023video} (5 Ens)& 0.3083 & 0.2189 & 0.2668 & 0.0636 & 0.0755 & 0.0355 & 0.3429 & 0.1343 \\
    Video-LLaMA (LLaMA-2 13B)~\cite{zhang2023video}  & 0.1186 & 0.2224 & 0.2006 & 0.0403 & 0.0802 & 0.0419 & 0.3143 & 0.0899 \\
    Video-LLaMA (LLaMA-2 7B)~\cite{zhang2023video} (5 Ens)  & 0.1458 & 0.1814 & 0.2281 & 0.0401 & 0.0835 & 0.0305 & 0.2286 & 0.1254 \\
    Video-LLaMA (LLaMA-2 7B)~\cite{zhang2023video} & 0.0235 & 0.1129 & 0.1125 & 0.0132 & 0.0462 & 0.0217 & 0.2571 & 0.0486 \\
    Valley (LLaMA-1 13B)~\cite{luo2023valley} (5 Ens) & 0.0000 & 0.0202 & 0.0213 & 0.0045 & 0.0000 & 0.0000 & 0.2571 & 0.0000 \\
    Valley (LLaMA-1 13B)~\cite{luo2023valley} & 0.0000 & 0.0109 & 0.0001 & 0.0000 & 0.0000 & 0.0000 & 0.1429 & 0.0000 \\
\hline
  \end{tabular}}
\caption{ {\bf Holistic (Counting) Experiment}: Comparing popular multi-modal LLMs across the JRDB-Social at three levels in {\bf accuracy} for the {\bf validation set}. BPC = Engagement of Body Position’s connection with the Content, SSC = Salient Scene Content, (5 Ens) = Five Ensemble Strategy.}
\label{table:HCV}
\end{table*}
\begin{table*}[b]
   \centering
   \footnotesize
   \scalebox{0.9}{
   \begin{tabular}{lccc|c|cccc}
    \toprule
    {\bf Multi-modal LLM} & \multicolumn{3}{c}{ {\bf Individual Level}} & \multicolumn{1}{c}{ {\bf Intra-Group Level }} & \multicolumn{4}{c}{{\bf Social Group Level}} \\
    \cline{2-9}
    & Gender & Age & Race & Interactions & BPC & SSC & Venue & Purpose \\
    \hline
    Video-LLaMA (LLaMA-2 13B)~\cite{zhang2023video} (5 Ens)& 0.2263 & 0.2239 & 0.2897 & 0.0894 & 0.0669 & 0.0093 & 0.4444 & 0.1581 \\
    Video-LLaMA (LLaMA-2 13B)~\cite{zhang2023video}  & 0.1142 & 0.2058 & 0.2109 & 0.0619 & 0.0822 & 0.0107 & 0.4000 & 0.1097 \\
    Video-LLaMA (LLaMA-2 7B)~\cite{zhang2023video} (5 Ens) & 0.1539 & 0.2019 & 0.2282 & 0.0680 & 0.0449 & 0.0065 & 0.3185 & 0.1122 \\
    Video-LLaMA (LLaMA-2 7B)~\cite{zhang2023video} & 0.0000 & 0.0076 & 0.0040 & 0.0036 & 0.0000 & 0.0000 & 0.3111 & 0.0078 \\
    Valley (LLaMA-1 13B)~\cite{luo2023valley} (5 Ens) & 0.0011 & 0.0229 & 0.0223 & 0.0000 & 0.0000 & 0.0000 & 0.2000 & 0.0007 \\
    Valley (LLaMA-1 13B)~\cite{luo2023valley} &  0.0011 & 0.0081 & 0.0038 & 0.0000 & 0.0010 & 0.0000 & 0.2222 & 0.0000 \\
\hline
  \end{tabular}}
\caption{{\bf Holistic (Counting) Experiment}: Comparing popular multi-modal LLMs across the JRDB-Social at three levels in {\bf accuracy} for the {\bf test set}. BPC = Engagement of Body Position’s connection with the Content, SSC = Salient Scene Content, (5 Ens) = Five Ensemble Strategy.}
\label{table:HCTe}
\end{table*}
\begin{table*}[b]
   \centering
   \footnotesize
   \scalebox{0.9}{
   \begin{tabular}{lccc|c|cccc}
    \toprule
    {\bf Multi-modal LLM} & \multicolumn{3}{c}{ {\bf Individual Level}} & \multicolumn{1}{c}{ {\bf Intra-Group Level }} & \multicolumn{4}{c}{{\bf Social Group Level}} \\
    \cline{2-9}
    & Gender & Age & Race & Interactions & BPC & SSC & Venue & Purpose \\
    \hline
    Video-LLaMA (LLaMA-2 13B)~\cite{zhang2023video} (5 Ens)& 0.9350 & 0.3320 & 0.5800 & 0.2604 & 0.1466 & 0.0755 & 0.1583& 0.3272 \\
    Video-LLaMA (LLaMA-2 13B)~\cite{zhang2023video}  & 0.9300 &0.3380& 0.5800 & 0.3466 &0.1983& 0.1283 & 0.1583&0.3490 \\
    Video-LLaMA (LLaMA-2 7B)~\cite{zhang2023video} (5 Ens) & 0.9350 & 0.3320 &0.5800& 0.1757 & 0.2572 & 0.0504 & 0.1583 & 0.2572 \\
    Video-LLaMA (LLaMA-2 7B)~\cite{zhang2023video} &0.9350  & 0.3320 & 0.5800 & 0.1938 & 0.1127 &0.0512& 0.1583 & 0.2654 \\
    Valley (LLaMA-1 13B)~\cite{luo2023valley} (5 Ens) & 0.8600 & 0.4940 & 0.5775 & 0.3576 & 0.6055 & 0.4459 & 0.1716 & 0.6090 \\
    Valley (LLaMA-1 13B)~\cite{luo2023valley} &  0.7700& 0.5320 &0.5825  &  0.3461&0.5666&0.4581&0.1666&0.6136 \\
\hline
  \end{tabular}}
\caption{ {\bf Holistic (Binary) Experiment}: Comparing popular multi-modal LLMs across the JRDB-Social at three levels in {\bf accuracy} for the {\bf train set}. BPC = Engagement of Body Position’s connection with the Content, SSC = Salient Scene Content, (5 Ens) = Five Ensemble Strategy.}
\label{table:HBTr}
\end{table*}
\begin{table*}[b]
   \centering
   \footnotesize
   \scalebox{0.9}{
   \begin{tabular}{lccc|c|cccc}
    \toprule
    {\bf Multi-modal LLM} & \multicolumn{3}{c}{ {\bf Individual Level}} & \multicolumn{1}{c}{ {\bf Intra-Group Level }} & \multicolumn{4}{c}{{\bf Social Group Level}} \\
    \cline{2-9}
    & Gender & Age & Race & Interactions & BPC & SSC & Venue & Purpose \\
    \hline
    Video-LLaMA (LLaMA-2 13B)~\cite{zhang2023video} (5 Ens)& 0.9714 & 0.3771 & 0.5571 & 0.3007  & 0.2111 & 0.1155 & 0.1667 & 0.3688 \\
    Video-LLaMA (LLaMA-2 13B)~\cite{zhang2023video}  & 0.9714 & 0.3886 & 0.5500 & 0.3633 &  0.2698 & 0.1878 & 0.1667 & 0.4520 \\
    Video-LLaMA (LLaMA-2 7B)~\cite{zhang2023video} (5 Ens) & 0.9714& 0.3771 & 0.5571 & 0.1918 & 0.1587 & 0.0623 & 0.1666 & 0.3168 \\
    Video-LLaMA (LLaMA-2 7B)~\cite{zhang2023video} & 0.9714 & 0.3771 & 0.5571 & 0.1986 & 0.1587 & 0.0629 & 0.1666 & 0.3194 \\
    Valley (LLaMA-1 13B)~\cite{luo2023valley} (5 Ens) &0.8857& 0.5029 & 0.5643 &  0.3415& 0.5683 & 0.4752 & 0.1762 & 0.6286 \\
    Valley (LLaMA-1 13B)~\cite{luo2023valley} & 0.7143 & 0.5086 & 0.5643 & 0.3633 & 0.5730 &0.4647&  0.2000& 0.6182 \\
\hline
  \end{tabular}}
\caption{ {\bf Holistic (Binary) Experiment}: Comparing popular multi-modal LLMs across the JRDB-Social at three levels in {\bf accuracy} for the {\bf validation set}. BPC = Engagement of Body Position’s connection with the Content, SSC = Salient Scene Content, (5 Ens) = Five Ensemble Strategy.}
\label{table:HBV}
\end{table*}
\begin{table*}[b]
   \centering
   \footnotesize
   \scalebox{0.9}{
   \begin{tabular}{lccc|c|cccc}
    \toprule
    {\bf Multi-modal LLM} & \multicolumn{3}{c}{ {\bf Individual Level}} & \multicolumn{1}{c}{ {\bf Intra-Group Level }} & \multicolumn{4}{c}{{\bf Social Group Level}} \\
    \cline{2-9}
    & Gender & Age & Race & Interactions & BPC & SSC & Venue & Purpose \\
    \hline
    Video-LLaMA (LLaMA-2 13B)~\cite{zhang2023video} (5 Ens)& 0.9815 & 0.4459 & 0.6185 & 0.3083 & 0.1621 & 0.0584 & 0.1667 & 0.3953 \\
    Video-LLaMA (LLaMA-2 13B)~\cite{zhang2023video}  & 0.9815 & 0.4533 & 0.6204 & 0.3746  & 0.2202 & 0.1306 & 0.1691 & 0.4249 \\
    Video-LLaMA (LLaMA-2 7B)~\cite{zhang2023video} (5 Ens) & 0.9815 & 0.4444 & 0.6222 & 0.2120 & 0.1210 & 0.0230 & 0.1667 & 0.3354 \\
    Video-LLaMA (LLaMA-2 7B)~\cite{zhang2023video} & 0.9778 & 0.4444 & 0.6222 & 0.2321 & 0.1222 & 0.0239 & 0.1667 & 0.3428 \\
    Valley (LLaMA-1 13B)~\cite{luo2023valley} (5 Ens) & 0.8704 & 0.5170 & 0.6259 & 0.3951 & 0.6235 & 0.5051 & 0.1815 & 0.6182 \\
    Valley (LLaMA-1 13B)~\cite{luo2023valley} & 0.8000 & 0.5170 & 0.6241 & 0.3898 & 0.6160 & 0.5046 & 0.1901 & 0.6108 \\
\hline
  \end{tabular}}
\caption{ {\bf Holistic (Binary) Experiment}: Comparing popular multi-modal LLMs across the JRDB-Social at three levels in {\bf accuracy} for the {\bf test set}. BPC = Engagement of Body Position’s connection with the Content, SSC = Salient Scene Content, (5 Ens) = Five Ensemble Strategy.}
\label{table:HBTe}
\end{table*}
\begin{table*}[!ht]
\centering
\footnotesize
\scalebox{0.9}{
\begin{tabular}{cc|ccccc|ccccc}
\hline
\multicolumn{2}{c}{} & \multicolumn{5}{c}{\textbf{Accuracy}} & \multicolumn{5}{c}{\textbf{F1 Score}} \\
\hline
\textbf{Resize} & \textbf{Pad Scale} & \textbf{Venue} & \textbf{Purpose} & \textbf{BPC} & \textbf{SSC} & \textbf{Avg.} & \textbf{Venue} & \textbf{Purpose} & \textbf{BPC} & \textbf{SSC} & \textbf{Avg.} \\
\hline
 & 1.0 & 0.1148 & 0.8199 & 0.5227 & 0.7322 & 0.5474 & 0.0686 & 0.2357 & 0.1818 & 0.0671 & 0.1383 \\
 & 1.2 & 0.0868 & 0.8211 & 0.5657 & 0.7452 & 0.5547 & 0.0553 & 0.2234 & 0.1997 & 0.0677 & 0.1365 \\
 & 1.4 & 0.0952 & 0.8173 & 0.5657 & 0.7195 & 0.5494 & 0.0664 & 0.2058 & 0.1761 & 0.0603 & 0.1271 \\
Frame-level & 1.6 & 0.0812 & 0.8290 & 0.5884 & 0.7801 & 0.5697 & 0.0420 & 0.2272 & 0.2027 & 0.0757 & 0.1369 \\
 & 1.8 & 0.0700 & 0.8196 & 0.6086 & 0.7455 & 0.5609 & 0.0335 & 0.2258 & 0.2003 & 0.0729 & 0.1331\\
 & 2.0 & 0.0868 & 0.8214 & 0.6162 & 0.7731 & 0.5744 & 0.0456 & 0.2236 & 0.1693 & 0.0706 & 0.1273\\
\hline
\end{tabular}}
\caption{Exploring diverse cropping scales at the group level: The left side of the table presents results in {\bf accuracy}, while the right side illustrates results in F1 score.}
\label{table:scale}
\end{table*}




%% file: sec_sup/Y_prompts.tex
\begin{prompt}[title={Prompt 1: Guided Perception Experiment},float*,width=\textwidth,floatplacement=t,breakable]\label{prompt:promptGP}
You are able to understand the visual content that the user provides. Follow the instructions carefully.\\
\textbf{\normalsize\# Gender} \\
What is the gender of the person in the centre of the video? Your answer should be one of \{gender categories (example: female)\}. Please think and generate only one word as the answer.\\
\textbf{\normalsize\# Age} \\
What is the age of the person in the centre of the video? Your answer should be one of \{age categories (example: middle adulthood)\}. Please think and generate only one word as the answer. \\
\textbf{\normalsize\# Race} \\
What is the race of the person in the centre of the video? Your answer should be one of \{race categories (example: Caucasian)\}. Please think and generate only one word as the answer. \\
\textbf{\normalsize\# Interaction} \\
What are the interactions between the people in the video? Your answer should be one or multiple of the following: \{interactions categories\}. Please think and list all possible answers. \\
\textbf{\normalsize\# BPC} \\
Where are the locations of most of the individuals in the group in the video? Your answer should be one or multiple of the following: \{BPC categories\}. Please think and generate only one word as the answer. \\
\textbf{\normalsize\# SSC} \\
What are the objects situated close to the group in the video? Your answer should be one or multiple of the following: \{SSC categories\}. Please think and list all possible answers. \\
\textbf{\normalsize\# Venue} \\
What is the venue of the groups of people in the video? Your answer should be one of the following: \{venue categories\}. Please think and generate only one word as the answer. \\
\textbf{\normalsize\# Purpose} \\
What are the aims and purposes of the group of people in the video? Your answer should be one or multiple of the following: \{purpose categories\}
\end{prompt}

\begin{prompt}[title={Prompt 2: Holistic Experiment (Counting Approach)},float*,width=\textwidth,floatplacement=t,breakable]\label{prompt:promptHC}
You are able to understand the visual content that the user provides. Follow the instructions carefully. \\
\textbf{\normalsize\# Gender} \\
How many \{gender categories (example: female)\} are in the video? Your answer should be number. Please think and generate only the number as the answer. \\
\textbf{\normalsize\# Age} \\
How many \{age categories (example: middle adulthood)\} are in the video? Your answer should be number. Please think and generate only the number as the answer.\\
\textbf{\normalsize\# Race} \\
How many \{race categories (example: Caucasian)\} are in the video? Your answer should be a number. Please think and generate only the number as the answer.\\
\textbf{\normalsize\# Interaction} \\
How many pairs of people are \{interaction categories\}? Your answer should be number. Please think and generate only the number as the answer. \\
\textbf{\normalsize\# BPC} \\
How many groups of people located on \{BPC categories (example: platform)\}? Your answer should be number. Please think and generate only the number as the answer. \\
\textbf{\normalsize\# SSC} \\
How many groups of people near the \{SSC categories (example: pillar)\}? Your answer should be number. Please think and generate only the
number as the answer.\\
\textbf{\normalsize\# Purpose} \\
How many groups of people are \{purpose categories (example: working)\}? Your answer should be number. Please think and generate only the number as the answer.
\end{prompt}

\begin{prompt}[title={Prompt 3: Holistic Experiment (Binary Approach)},float*,width=\textwidth,floatplacement=t,breakable]\label{prompt:promptHB}
You are able to understand the visual content that the user provides. Follow the instructions carefully. \\
\textbf{\normalsize\# Gender} \\
Do you see \{gender categories (example: female)\} in the video? Your answer should be yes or no. Please think and generate only the word as the answer.\\
\textbf{\normalsize\# Age} \\
Do you see \{age categories (example: middle adulthood)\} in the video? Your answer should be yes or no. Please think and generate only the word as the answer.\\
\textbf{\normalsize\# Race} \\
Do you see \{race categories (example: Caucasian)\} in the video? Your answer should be yes or no. Please think and generate only the word as the answer.\\
\textbf{\normalsize\# Interaction} \\
Do you see any pair of people are \{interaction categories (example: standing together)\}? Your answer should be yes or no. Please think and generate only the word as the answer.\\
\textbf{\normalsize\# BPC} \\
Do you see any group located on \{BPC categories (example: floor)\}? Your answer should be yes or no. Please think and generate only the word as the answer.\\
\textbf{\normalsize\# SSC} \\
Do you see any group near the \{SSC categories (example: pillar)\}? Your answer should be yes or no. Please think and generate only the word as the answer.\\
\textbf{\normalsize\# Purpose} \\
Do you see any group are \{purpose categories (example: socializing)\}? Your answer should be yes or no. Please think and generate only the word as the answer.
\end{prompt}

%% file: main.bbl
\begin{thebibliography}{52}
\providecommand{\natexlab}[1]{#1}
\providecommand{\url}[1]{\texttt{#1}}
\expandafter\ifx\csname urlstyle\endcsname\relax
  \providecommand{\doi}[1]{doi: #1}\else
  \providecommand{\doi}{doi: \begingroup \urlstyle{rm}\Url}\fi

\bibitem[Ehsanpour et~al.(2020)Ehsanpour, Abedin, Saleh, Shi, Reid, and Rezatofighi]{ehsanpour2020joint}
Mahsa Ehsanpour, Alireza Abedin, Fatemeh Saleh, Javen Shi, Ian Reid, and Hamid Rezatofighi.
\newblock Joint learning of social groups, individuals action and sub-group activities in videos.
\newblock In \emph{Computer Vision--ECCV 2020: 16th European Conference, Glasgow, UK, August 23--28, 2020, Proceedings, Part IX 16}, pages 177--195. Springer, 2020.

\bibitem[Martin-Martin et~al.(2021)Martin-Martin, Patel, Rezatofighi, Shenoi, Gwak, Frankel, Sadeghian, and Savarese]{martin2021jrdb}
Roberto Martin-Martin, Mihir Patel, Hamid Rezatofighi, Abhijeet Shenoi, JunYoung Gwak, Eric Frankel, Amir Sadeghian, and Silvio Savarese.
\newblock Jrdb: A dataset and benchmark of egocentric robot visual perception of humans in built environments.
\newblock \emph{IEEE transactions on pattern analysis and machine intelligence}, 2021.

\bibitem[Gong et~al.(2011)Gong, Loy, and Xiang]{gong2011security}
Shaogang Gong, Chen~Change Loy, and Tao Xiang.
\newblock Security and surveillance.
\newblock \emph{Visual analysis of humans: Looking at people}, pages 455--472, 2011.

\bibitem[Barua et~al.(2020)Barua, Sarkar, Kumar, Pal, et~al.]{barua2020can}
Hrishav~Bakul Barua, Chayan Sarkar, Achanna~Anil Kumar, Arpan Pal, et~al.
\newblock I can attend a meeting too! towards a human-like telepresence avatar robot to attend meeting on your behalf.
\newblock \emph{arXiv preprint arXiv:2006.15647}, 2020.

\bibitem[Hanheide et~al.(2017)Hanheide, Hebesberger, and Krajn{\'\i}k]{hanheide2017and}
Marc Hanheide, Denise Hebesberger, and Tom{\'a}{\v{s}} Krajn{\'\i}k.
\newblock The when, where, and how: An adaptive robotic info-terminal for care home residents.
\newblock In \emph{Proceedings of the 2017 ACM/IEEE International Conference on Human-Robot Interaction}, pages 341--349, 2017.

\bibitem[Logan et~al.(2019)Logan, Breazeal, Goodwin, Jeong, O’Connell, Smith-Freedman, Heathers, and Weinstock]{logan2019social}
Deirdre~E Logan, Cynthia Breazeal, Matthew~S Goodwin, Sooyeon Jeong, Brianna O’Connell, Duncan Smith-Freedman, James Heathers, and Peter Weinstock.
\newblock Social robots for hospitalized children.
\newblock \emph{Pediatrics}, 144\penalty0 (1), 2019.

\bibitem[Yatskar et~al.(2016)Yatskar, Zettlemoyer, and Farhadi]{yatskar2016situation}
Mark Yatskar, Luke Zettlemoyer, and Ali Farhadi.
\newblock Situation recognition: Visual semantic role labeling for image understanding.
\newblock In \emph{Proceedings of the IEEE conference on computer vision and pattern recognition}, pages 5534--5542, 2016.

\bibitem[Tanisik et~al.(2016)Tanisik, Zalluhoglu, and Ikizler-Cinbis]{tanisik2016facial}
Gokhan Tanisik, Cemil Zalluhoglu, and Nazli Ikizler-Cinbis.
\newblock Facial descriptors for human interaction recognition in still images.
\newblock \emph{Pattern Recognition Letters}, 73:\penalty0 44--51, 2016.

\bibitem[Tanisik et~al.(2021)Tanisik, Zalluhoglu, and Ikizler-Cinbis]{tanisik2021multi}
Gokhan Tanisik, Cemil Zalluhoglu, and Nazli Ikizler-Cinbis.
\newblock Multi-stream pose convolutional neural networks for human interaction recognition in images.
\newblock \emph{Signal Processing: Image Communication}, 95:\penalty0 116265, 2021.

\bibitem[Ronchi and Perona(2015)]{ronchi2015describing}
Matteo~Ruggero Ronchi and Pietro Perona.
\newblock Describing common human visual actions in images.
\newblock \emph{arXiv preprint arXiv:1506.02203}, 2015.

\bibitem[Ryoo and Aggarwal(2010)]{ryoo2010ut}
Michael~S Ryoo and JK~Aggarwal.
\newblock Ut-interaction dataset, icpr contest on semantic description of human activities (sdha).
\newblock In \emph{IEEE International Conference on Pattern Recognition Workshops}, volume~2, page~4, 2010.

\bibitem[Lee and Lee(2022)]{lee2022human}
Dong-Gyu Lee and Seong-Whan Lee.
\newblock Human interaction recognition framework based on interacting body part attention.
\newblock \emph{Pattern Recognition}, 128:\penalty0 108645, 2022.

\bibitem[Patron-Perez et~al.(2010)Patron-Perez, Marszalek, Zisserman, and Reid]{patron2010high}
Alonso Patron-Perez, Marcin Marszalek, Andrew Zisserman, and Ian Reid.
\newblock High five: Recognising human interactions in tv shows.
\newblock In \emph{BMVC}, volume~1, page~33, 2010.

\bibitem[Marszalek et~al.(2009)Marszalek, Laptev, and Schmid]{marszalek2009actions}
Marcin Marszalek, Ivan Laptev, and Cordelia Schmid.
\newblock Actions in context.
\newblock In \emph{2009 IEEE Conference on Computer Vision and Pattern Recognition}, pages 2929--2936. IEEE, 2009.

\bibitem[Alameda-Pineda et~al.(2015)Alameda-Pineda, Staiano, Subramanian, Batrinca, Ricci, Lepri, Lanz, and Sebe]{alameda2015salsa}
Xavier Alameda-Pineda, Jacopo Staiano, Ramanathan Subramanian, Ligia Batrinca, Elisa Ricci, Bruno Lepri, Oswald Lanz, and Nicu Sebe.
\newblock Salsa: A novel dataset for multimodal group behavior analysis.
\newblock \emph{IEEE transactions on pattern analysis and machine intelligence}, 38\penalty0 (8):\penalty0 1707--1720, 2015.

\bibitem[Wang et~al.(2020)Wang, Zhang, Zhu, Guo, Yuan, Xiang, Wang, Ding, Brady, Dai, et~al.]{wang2020panda}
Xueyang Wang, Xiya Zhang, Yinheng Zhu, Yuchen Guo, Xiaoyun Yuan, Liuyu Xiang, Zerun Wang, Guiguang Ding, David Brady, Qionghai Dai, et~al.
\newblock Panda: A gigapixel-level human-centric video dataset.
\newblock In \emph{Proceedings of the IEEE/CVF conference on computer vision and pattern recognition}, pages 3268--3278, 2020.

\bibitem[Ehsanpour et~al.(2022)Ehsanpour, Saleh, Savarese, Reid, and Rezatofighi]{ehsanpour2022jrdb}
Mahsa Ehsanpour, Fatemeh Saleh, Silvio Savarese, Ian Reid, and Hamid Rezatofighi.
\newblock Jrdb-act: A large-scale dataset for spatio-temporal action, social group and activity detection.
\newblock In \emph{Proceedings of the IEEE/CVF Conference on Computer Vision and Pattern Recognition}, pages 20983--20992, 2022.

\bibitem[Krishna et~al.(2017)Krishna, Hata, Ren, Fei-Fei, and Carlos~Niebles]{krishna2017dense}
Ranjay Krishna, Kenji Hata, Frederic Ren, Li~Fei-Fei, and Juan Carlos~Niebles.
\newblock Dense-captioning events in videos.
\newblock In \emph{Proceedings of the IEEE international conference on computer vision}, pages 706--715, 2017.

\bibitem[Huang et~al.(2020)Huang, Pang, Zhu, Rivera, and Soricut]{huang2020multimodal}
Gabriel Huang, Bo~Pang, Zhenhai Zhu, Clara Rivera, and Radu Soricut.
\newblock Multimodal pretraining for dense video captioning.
\newblock In \emph{AACL-IJCNLP 2020}, 2020.

\bibitem[Zhou et~al.(2018)Zhou, Xu, and Corso]{zhou2018towards}
Luowei Zhou, Chenliang Xu, and Jason Corso.
\newblock Towards automatic learning of procedures from web instructional videos.
\newblock In \emph{Proceedings of the AAAI Conference on Artificial Intelligence}, volume~32, 2018.

\bibitem[Tan et~al.(2020)Tan, Liu, Wang, and Zha]{tan2020learning}
Ganchao Tan, Daqing Liu, Meng Wang, and Zheng-Jun Zha.
\newblock Learning to discretely compose reasoning module networks for video captioning.
\newblock \emph{arXiv preprint arXiv:2007.09049}, 2020.

\bibitem[Wang et~al.(2019)Wang, Wu, Chen, Li, Wang, and Wang]{wang2019vatex}
Xin Wang, Jiawei Wu, Junkun Chen, Lei Li, Yuan-Fang Wang, and William~Yang Wang.
\newblock Vatex: A large-scale, high-quality multilingual dataset for video-and-language research.
\newblock In \emph{Proceedings of the IEEE/CVF International Conference on Computer Vision}, pages 4581--4591, 2019.

\bibitem[Vendrow et~al.(2023)Vendrow, Le, Cai, and Rezatofighi]{vendrow2023jrdb}
Edward Vendrow, Duy~Tho Le, Jianfei Cai, and Hamid Rezatofighi.
\newblock Jrdb-pose: A large-scale dataset for multi-person pose estimation and tracking.
\newblock In \emph{Proceedings of the IEEE/CVF Conference on Computer Vision and Pattern Recognition}, pages 4811--4820, 2023.

\bibitem[Zhang et~al.(2023)Zhang, Li, and Bing]{zhang2023video}
Hang Zhang, Xin Li, and Lidong Bing.
\newblock Video-llama: An instruction-tuned audio-visual language model for video understanding.
\newblock \emph{arXiv preprint arXiv:2306.02858}, 2023.

\bibitem[Wu et~al.(2023{\natexlab{a}})Wu, Fei, Qu, Ji, and Chua]{wu2023next}
Shengqiong Wu, Hao Fei, Leigang Qu, Wei Ji, and Tat-Seng Chua.
\newblock Next-gpt: Any-to-any multimodal llm.
\newblock \emph{arXiv preprint arXiv:2309.05519}, 2023{\natexlab{a}}.

\bibitem[Maaz et~al.(2023)Maaz, Rasheed, Khan, and Khan]{maaz2023video}
Muhammad Maaz, Hanoona Rasheed, Salman Khan, and Fahad~Shahbaz Khan.
\newblock Video-chatgpt: Towards detailed video understanding via large vision and language models.
\newblock \emph{arXiv preprint arXiv:2306.05424}, 2023.

\bibitem[Li et~al.(2023{\natexlab{a}})Li, He, Wang, Li, Wang, Luo, Wang, Wang, and Qiao]{li2023videochat}
KunChang Li, Yinan He, Yi~Wang, Yizhuo Li, Wenhai Wang, Ping Luo, Yali Wang, Limin Wang, and Yu~Qiao.
\newblock Videochat: Chat-centric video understanding.
\newblock \emph{arXiv preprint arXiv:2305.06355}, 2023{\natexlab{a}}.

\bibitem[Vicol et~al.(2018)Vicol, Tapaswi, Castrejon, and Fidler]{vicol2018moviegraphs}
Paul Vicol, Makarand Tapaswi, Lluis Castrejon, and Sanja Fidler.
\newblock Moviegraphs: Towards understanding human-centric situations from videos.
\newblock In \emph{Proceedings of the IEEE conference on computer vision and pattern recognition}, pages 8581--8590, 2018.

\bibitem[Rasouli et~al.(2017)Rasouli, Kotseruba, and Tsotsos]{rasouli2017ICCVW}
Amir Rasouli, Iuliia Kotseruba, and John~K Tsotsos.
\newblock Are they going to cross? a benchmark dataset and baseline for pedestrian crosswalk behavior.
\newblock In \emph{Proceedings of the IEEE International Conference on Computer Vision Workshops}, pages 206--213, 2017.

\bibitem[Rasouli et~al.(2019)Rasouli, Kotseruba, Kunic, and Tsotsos]{rasouli2019pie}
Amir Rasouli, Iuliia Kotseruba, Toni Kunic, and John~K Tsotsos.
\newblock Pie: A large-scale dataset and models for pedestrian intention estimation and trajectory prediction.
\newblock In \emph{Proceedings of the IEEE/CVF International Conference on Computer Vision}, pages 6262--6271, 2019.

\bibitem[Gu et~al.(2018)Gu, Sun, Ross, Vondrick, Pantofaru, Li, Vijayanarasimhan, Toderici, Ricco, Sukthankar, et~al.]{gu2018ava}
Chunhui Gu, Chen Sun, David~A Ross, Carl Vondrick, Caroline Pantofaru, Yeqing Li, Sudheendra Vijayanarasimhan, George Toderici, Susanna Ricco, Rahul Sukthankar, et~al.
\newblock Ava: A video dataset of spatio-temporally localized atomic visual actions.
\newblock In \emph{Proceedings of the IEEE conference on computer vision and pattern recognition}, pages 6047--6056, 2018.

\bibitem[Yun et~al.(2012)Yun, Honorio, Chattopadhyay, Berg, and Samaras]{yun2012two}
Kiwon Yun, Jean Honorio, Debaleena Chattopadhyay, Tamara~L Berg, and Dimitris Samaras.
\newblock Two-person interaction detection using body-pose features and multiple instance learning.
\newblock In \emph{2012 IEEE computer society conference on computer vision and pattern recognition workshops}, pages 28--35. IEEE, 2012.

\bibitem[Shahroudy et~al.(2016)Shahroudy, Liu, Ng, and Wang]{shahroudy2016ntu}
Amir Shahroudy, Jun Liu, Tian-Tsong Ng, and Gang Wang.
\newblock Ntu rgb+ d: A large scale dataset for 3d human activity analysis.
\newblock In \emph{Proceedings of the IEEE conference on computer vision and pattern recognition}, pages 1010--1019, 2016.

\bibitem[Piergiovanni and Ryoo(2018)]{mlbcaptions2018}
AJ~Piergiovanni and Michael~S. Ryoo.
\newblock Learning shared multimodal embeddings with unpaired data.
\newblock \emph{arXiv preprint arXiv:1806.08251}, 2018.

\bibitem[Luo et~al.(2020)Luo, Ye, Adams, Li, Newman, and Wang]{luo2020arbee}
Yu~Luo, Jianbo Ye, Reginald~B Adams, Jia Li, Michelle~G Newman, and James~Z Wang.
\newblock Arbee: Towards automated recognition of bodily expression of emotion in the wild.
\newblock \emph{International journal of computer vision}, 128:\penalty0 1--25, 2020.

\bibitem[Cui et~al.(2021)Cui, Khandelwal, Artzi, Snavely, and Averbuch-Elor]{cui2021s}
Yuqing Cui, Apoorv Khandelwal, Yoav Artzi, Noah Snavely, and Hadar Averbuch-Elor.
\newblock Who's waldo? linking people across text and images.
\newblock In \emph{Proceedings of the IEEE/CVF International Conference on Computer Vision}, pages 1374--1384, 2021.

\bibitem[Orcesi et~al.(2021)Orcesi, Audigier, Toukam, and Luvison]{orcesi2021detecting}
Astrid Orcesi, Romaric Audigier, Fritz~Poka Toukam, and Bertrand Luvison.
\newblock Detecting human-to-human-or-object (h 2 o) interactions with diabolo.
\newblock In \emph{2021 16th IEEE International Conference on Automatic Face and Gesture Recognition (FG 2021)}, pages 1--8. IEEE, 2021.

\bibitem[Chowdhery et~al.(2022)Chowdhery, Narang, Devlin, Bosma, Mishra, Roberts, Barham, Chung, Sutton, Gehrmann, et~al.]{chowdhery2022palm}
Aakanksha Chowdhery, Sharan Narang, Jacob Devlin, Maarten Bosma, Gaurav Mishra, Adam Roberts, Paul Barham, Hyung~Won Chung, Charles Sutton, Sebastian Gehrmann, et~al.
\newblock Palm: Scaling language modeling with pathways.
\newblock \emph{arXiv preprint arXiv:2204.02311}, 2022.

\bibitem[Bai et~al.(2022)Bai, Kadavath, Kundu, Askell, Kernion, Jones, Chen, Goldie, Mirhoseini, McKinnon, et~al.]{bai2022constitutional}
Yuntao Bai, Saurav Kadavath, Sandipan Kundu, Amanda Askell, Jackson Kernion, Andy Jones, Anna Chen, Anna Goldie, Azalia Mirhoseini, Cameron McKinnon, et~al.
\newblock Constitutional ai: Harmlessness from ai feedback.
\newblock \emph{arXiv preprint arXiv:2212.08073}, 2022.

\bibitem[Ouyang et~al.(2022)Ouyang, Wu, Jiang, Almeida, Wainwright, Mishkin, Zhang, Agarwal, Slama, Ray, et~al.]{ouyang2022training}
Long Ouyang, Jeffrey Wu, Xu~Jiang, Diogo Almeida, Carroll Wainwright, Pamela Mishkin, Chong Zhang, Sandhini Agarwal, Katarina Slama, Alex Ray, et~al.
\newblock Training language models to follow instructions with human feedback.
\newblock \emph{Advances in Neural Information Processing Systems}, 35:\penalty0 27730--27744, 2022.

\bibitem[Radford et~al.(2019)Radford, Wu, Child, Luan, Amodei, Sutskever, et~al.]{radford2019language}
Alec Radford, Jeffrey Wu, Rewon Child, David Luan, Dario Amodei, Ilya Sutskever, et~al.
\newblock Language models are unsupervised multitask learners.
\newblock \emph{OpenAI blog}, 1\penalty0 (8):\penalty0 9, 2019.

\bibitem[Wu et~al.(2023{\natexlab{b}})Wu, Yin, Qi, Wang, Tang, and Duan]{wu2023visual}
Chenfei Wu, Shengming Yin, Weizhen Qi, Xiaodong Wang, Zecheng Tang, and Nan Duan.
\newblock Visual chatgpt: Talking, drawing and editing with visual foundation models.
\newblock \emph{arXiv preprint arXiv:2303.04671}, 2023{\natexlab{b}}.

\bibitem[Luo et~al.(2023)Luo, Zhao, Yang, Dong, Qiu, Lu, Wang, and Wei]{luo2023valley}
Ruipu Luo, Ziwang Zhao, Min Yang, Junwei Dong, Minghui Qiu, Pengcheng Lu, Tao Wang, and Zhongyu Wei.
\newblock Valley: Video assistant with large language model enhanced ability.
\newblock \emph{arXiv preprint arXiv:2306.07207}, 2023.

\bibitem[Li et~al.(2023{\natexlab{b}})Li, Zhang, Chen, Wang, Yang, and Liu]{li2023otter}
Bo~Li, Yuanhan Zhang, Liangyu Chen, Jinghao Wang, Jingkang Yang, and Ziwei Liu.
\newblock Otter: A multi-modal model with in-context instruction tuning.
\newblock \emph{arXiv preprint arXiv:2305.03726}, 2023{\natexlab{b}}.

\bibitem[Sur{\'\i}s et~al.(2023)Sur{\'\i}s, Menon, and Vondrick]{suris2023vipergpt}
D{\'\i}dac Sur{\'\i}s, Sachit Menon, and Carl Vondrick.
\newblock Vipergpt: Visual inference via python execution for reasoning.
\newblock \emph{arXiv preprint arXiv:2303.08128}, 2023.

\bibitem[Zhu et~al.(2023)Zhu, Chen, Shen, Li, and Elhoseiny]{zhu2023minigpt}
Deyao Zhu, Jun Chen, Xiaoqian Shen, Xiang Li, and Mohamed Elhoseiny.
\newblock Minigpt-4: Enhancing vision-language understanding with advanced large language models.
\newblock \emph{arXiv preprint arXiv:2304.10592}, 2023.

\bibitem[Dai et~al.(2023)Dai, Li, Li, Tiong, Zhao, Wang, Li, Fung, and Hoi]{instructblip}
Wenliang Dai, Junnan Li, Dongxu Li, Anthony Meng~Huat Tiong, Junqi Zhao, Weisheng Wang, Boyang Li, Pascale Fung, and Steven Hoi.
\newblock Instructblip: Towards general-purpose vision-language models with instruction tuning, 2023.

\bibitem[Li et~al.(2023{\natexlab{c}})Li, Yin, Li, Chen, Wang, Ren, Li, Yang, Xu, Sun, et~al.]{li2023m}
Lei Li, Yuwei Yin, Shicheng Li, Liang Chen, Peiyi Wang, Shuhuai Ren, Mukai Li, Yazheng Yang, Jingjing Xu, Xu~Sun, et~al.
\newblock {$\text{M}^3$}it: A large-scale dataset towards multi-modal multilingual instruction tuning.
\newblock \emph{arXiv preprint arXiv:2306.04387}, 2023{\natexlab{c}}.

\bibitem[Wang et~al.(2023)Wang, Chen, Chen, Wu, Zhu, Zeng, Luo, Lu, Zhou, Qiao, et~al.]{wang2023visionllm}
Wenhai Wang, Zhe Chen, Xiaokang Chen, Jiannan Wu, Xizhou Zhu, Gang Zeng, Ping Luo, Tong Lu, Jie Zhou, Yu~Qiao, et~al.
\newblock Visionllm: Large language model is also an open-ended decoder for vision-centric tasks.
\newblock \emph{arXiv preprint arXiv:2305.11175}, 2023.

\bibitem[Papineni et~al.(2002)Papineni, Roukos, Ward, and Zhu]{papineni2002bleu}
Kishore Papineni, Salim Roukos, Todd Ward, and Wei-Jing Zhu.
\newblock Bleu: a method for automatic evaluation of machine translation.
\newblock In \emph{Proceedings of the 40th annual meeting of the Association for Computational Linguistics}, pages 311--318, 2002.

\bibitem[Chin-Yew(2004)]{chin2004rouge}
Lin Chin-Yew.
\newblock Rouge: A package for automatic evaluation of summaries.
\newblock In \emph{Proceedings of the Workshop on Text Summarization Branches Out, 2004}, 2004.

\bibitem[Banerjee and Lavie(2005)]{banerjee2005meteor}
Satanjeev Banerjee and Alon Lavie.
\newblock Meteor: An automatic metric for mt evaluation with improved correlation with human judgments.
\newblock In \emph{Proceedings of the acl workshop on intrinsic and extrinsic evaluation measures for machine translation and/or summarization}, pages 65--72, 2005.

\end{thebibliography}
